\documentclass[review,3p,11pt,times]{elsarticle}
\usepackage{algorithm,algcompatible,amsmath}
\algnewcommand\INPUT{\item[\textbf{Input:}]}%
\algnewcommand\OUTPUT{\item[\textbf{Output:}]}%
\usepackage{units}
\usepackage{hyperref} %lineno, 
\usepackage{amsfonts}
\usepackage{amsmath}
\usepackage{amssymb}
\usepackage{xcolor}
\usepackage{graphicx}
\usepackage{subfig}
\usepackage{enumitem}
\usepackage{multirow}
\usepackage{physics}
\biboptions{sort&compress}


\journal {arXiv} %{Journal of Computational Physics} %{arXiv}

\begin{document}

\begin{frontmatter}
% ==============================================================================
%Title
% ==============================================================================

\title{\Large Understanding the role of autoencoders for stiff dynamical systems using information theory}
% ==============================================================================
%Author information
% ==============================================================================
\author[mymainaddress]{Vijayamanikandan Vijayarangan\corref{mycorrespondingauthor}}
\cortext[mycorrespondingauthor]{Corresponding author}
\ead{vijayamanikandan.vijayarangan@kaust.edu.sa}
%\ead[url]{www.elsevier.com}
\author[mymainaddress,mymainaddress2]{Harshavardhana A. Uranakara}
\author[mymainaddress]{Francisco E. Hern\'andez-P\'erez}
\author[mymainaddress]{Hong G. Im}

% ==============================================================================
% Author affiliation
% ==============================================================================
%
%
\address[mymainaddress]{Clean Energy Research Platform, Physical Science and Engineering (PSE) Division, \\ King Abdullah University of Science and Technology (KAUST), Thuwal 23955-6900, Saudi Arabia}
\address[mymainaddress2]{Ansys Software Pvt. Ltd., Bangalore, Karnataka, India}
%
% ==============================================================================
% Abstract
% ==============================================================================
\begin{abstract}
Using the information theory, this study provides insights into how the construction of latent space of autoencoder (AE) using deep neural network (DNN) training finds a smooth low-dimensional manifold in the stiff dynamical system. Our recent study\cite{vijayarangan2023data} reported that an autoencoder (AE) combined with neural ODE (NODE) as a surrogate reduced order model (ROM) for the integration of stiff chemically reacting systems led to a significant reduction in the temporal stiffness, and the behavior was attributed to the identification of a slow invariant manifold by the nonlinear projection of the AE. The present work offers fundamental understanding of the mechanism by employing concepts from information theory and better mixing. The learning mechanism of both the encoder and decoder are explained by plotting the evolution of mutual information and identifying two different phases. Subsequently, the density distribution is plotted for the physical and latent variables, which shows the transformation of the \emph{rare event} in the physical space to a \emph{highly likely} (more probable) event in the latent space provided by the nonlinear autoencoder. Finally, the nonlinear transformation leading to density redistribution is explained using concepts from information theory and probability.
\end{abstract}

% ==============================================================================
% Keywords
% ==============================================================================

\end{frontmatter}

%\linenumbers

% ==============================================================================
% Introduction
% ==============================================================================
\section{Introduction} \label{Sec:Introduction}

Many complex physical phenomena such as turbulent chemically-reacting flows are dynamical systems, which are mathematically described by time-dependent nonlinear partial differential equations (PDE). The equations are computationally solved in a discretized form by constructing a system of ordinary differential equations (ODE), and are integrated by the method of lines. The dimensionality of the ODE system is large, proportional to the number of solution variables and the number of computational cells. The large dimensionality is also accompanied by a wide range of time scales, posing a numerical stiffness that requires many small time steps and often forms the bottleneck. As such, the computational demand to simulate a modest physical size of reacting flows can easily reach tens of millions of core hours in modern high performance computing (HPC) hardware. As the physical problems under study become more sophisticated, effective acceleration of simulations by various means of reduced dimensionality and temporal stiffness without loss of fidelity is of paramount importance.

The key principle for developing high-fidelity reduced-order models (ROM) is to identify the correct minimal coordinates, referred to as intrinsic low-dimensional manifolds, in the hyper-dimensional phase space ~\cite{steinhauser2017computational,brunton2022data}. In general, the physical state variables are mapped onto the low-dimensional manifolds using a linear or nonlinear transformation, thereby reducing the number of physical state variables and the dynamic range of time scales. The successful development of ROMs allows significantly reduced computational costs and facilitates the physical interpretation of complex phenomena \cite{lucia2004reduced}. 

The development of ROMs can be broadly classified into two groups: physics-based and data-driven. In the physics-based ROM approach, the governing equations are simplified using assumptions deduced from the observed solution behavior. An earlier example is the modeling of atmospheric flows, where the Lorenz equations \cite{lorenz1963deterministic} described two-dimensional convective flows using three ODEs. In this approach, the state-space (Poincar\'e) model linearizes the system at fixed points and constructs models using averaging and homogenization methods~\cite{redkar2008direct, pavliotis2008multiscale}. For reactive systems, the computational singular perturbation (CSP) ~\cite{lam1989understanding,valorani2005chemical} has been used to identify low-dimensional manifolds to reduce the number of state variables and eliminate unnecessary fast time scales. This approach has been successfully implemented in ODE solvers for the accelerated integration of complex chemically reacting systems~\cite{galassi2022pycsp, galassi_CSPANN}.

With advances in data science and machine-learning algorithms, data-driven ROM development has recently been gaining strong research interests. In this approach, a large amount of generated data from experiments or simulations is further processed to determine a proper transformation map. The method can be broadly classified into (i) Koopman analysis and (ii) neural networks (NN). In the former, a nonlinear coordinate transformation maps a dynamical system to a linear system of observables \cite{bevanda2021koopman, brunton2022modern}. This has been applied to computational fluid dynamics (CFD) \cite{ito1998reduced, ravindran2000reduced, brunton2013reduced}, combined with matrix decomposition tools such as principal component analysis (PCA) and dynamic mode decomposition (DMD) \cite{schmid2022dynamic}. PCA has also been used successfully in chemically reacting flow simulations to reduce state variables~\cite{ malik2023dimensionality, malik2024combined}. In the latter approach based on NN, the input data is processed using a series of affine transformations along with the nonlinear activation functions to construct a ROM. This approach has widely been used in CFD for model discovery \cite{ling2016reynolds, zhang2020data}, flow optimization and control \cite{verma2018efficient, vignon2023effective}, and reconstruction of high-resolution flow fields from sparse measurements \cite{fukami2023super, kim2021unsupervised, guemes2022super, yousif2021high, vijayarangan2024reconstruction}.

A variation of the standard deep neural network (DNN) approach for dimensionality reduction is the autoencoder (AE), which maps the physical state variables into the low-dimensional manifolds, referred to as the latent space. Unlike Koopman analysis, AEs offer the flexibility of using either linear or nonlinear transformations to obtain the latent space variables. The reduced variables in the latent space are approximated using time-series forecasting methods (e.g. transformers, neural ODE, recurrent neural networks, etc.), which are recovered back in physical space using a decoder. This three-stage approach is widely adopted in the study of dynamical systems, with notable improvements in ROM fidelity along with computational speedup \cite{champion2019data, dikemanNODE, lee2021parameterized, vijayarangan2023data}. In our recent study~\cite{vijayarangan2023data}, the AE combined with neural ODE (NODE) was used in a homogeneous reactive system with a large number of reactive scalars, and a significant level of stiffness reduction was achieved through an optimal training strategy to dynamics-informed construction of the latent space. 

Although AE-based methods have been shown to be an effective tool for developing ROMs with compressed dimensionality and dynamics, the optimal design of AE is based on many fine-tunings of the hyper-parameters, such as the activation function, number of units per layer, the total number of hidden layers, dimension of the latent space, often in a trial-and-error manner. Moreover, there is little understanding of the underlying mathematical or physical explanation of how training leads to finding an optimal latent space with a removal of temporal stiffness. 

To provide insights into this issue, several mathematical frameworks have been used to understand the mechanism and behavior of the ``black box" of neural networks in representing the complex input data. 
%These approaches can be classified as: (i) functional analysis-based -- approximation theory \cite{augustine2024survey}; (ii) geometry-based -- non-Euclidean geometric deep learning \cite{kratsios2022universal}; (iii) algebra-based -- nonlinear projection method \cite{principe2015universal}. (iv) probability and statistics-based -- information bottleneck theory \cite{shwartz2017opening, alemi2016deep} and deep mixing theory \cite{bengio2013better}. 
For example, the universal approximation theorem (UAT) was used to determine the validity of approximating any continuous function using NN~\cite{augustine2024survey}. Differential geometry concepts have also been used to understand the non-Euclidean nature of the data learning framework~\cite{kratsios2022universal}. Alternatively, linear algebra-based projection theorems were used to explain the process of projection on linear sub-manifolds~\cite{principe2015universal}, although it was argued that the linear theory is limited to shallow NNs only~\cite{liu2020selection}. A combination of different mathematical tools was also used to explain the NN-based data learning process. For example, neural tangent kernel theory \cite{paccolat2021geometric} used concepts from linear algebra, functional analysis, geometry, and probability. Similarly, Liu and Markowich \cite{liu2020selection} proposed a PDE-based analogy for NN using the concepts of functional analysis, variational calculus, and optimal control. Differing from these research works, the present study employs a statistical method, specifically the information theory, that has been gaining popularity in explaining the learning process of NN.

%For probabilistic consideration of DNN, the conditional and marginal probabilities are learned in supervised learning, provided that data represents the joint probability of the inputs and outputs. In information theory, mutual information (MI) is a metric to quantify the relevance between two random variables (in this case input and output variables). Principe et al.~\cite{principe1999introduction} proposed an NN optimization by exploring the MI-based data learning process. This subject will be discussed in detail in section \ref{Sec:Elements_Info_theory}.  %These methods were also demonstrated successfully in independent component analysis and blind source separation applications.

%Tishby et. al. \cite{tishby2000information} introduced a variational principle-based iterative algorithm for the extraction of relevant information. Until that point, the traditional data-driven approaches used mean square error-based criteria. Renyi's entropy definition is used in the approach to measure the probability density function for MI instead of iterative-based algorithms.
%To begin with, information theory is primarily used in communication theory \cite{cover1999elements} to study the maximum limit of data compression and transmission rate of communication.

Statistics-based methods have been widely employed to understand the working principles and learning mechanisms of the NN. The theoretical foundations of information bottleneck theory (IBT) were given by Tishby et al.~ \cite{tishby2000information},  introducing the concept of ``relevant information'' in a signal and providing an iterative algorithm to extract ``mutual information (MI)." The application of IBT opened a new pathway to understanding NN, where a qualitative information plane was introduced to show MI between the hidden layer, the input, and the output variables during DNN training. The phase transition in the information curve was linked to properties of the optimal DNN architecture and information trade-off within the layers. Schwartz and Tishby \cite{shwartz2017opening} subsequently extended the IBT to understand a practical DNN, employing a fully connected feedforward network with seven hidden layers consisting of 12-10-7-5-4-3-2 neurons with the hyperbolic tangent or sigmoid activation functions. Using the training data set consisting of spherically symmetric binary decision rules, the DNN training process was classified as an initial fitting phase followed by a comprehensive compression phase, where the latter plays the role of generalization through diffusion-like behavior.
%made the following three specific claims based on the information plane: (i) DNN undergoes two distinct phases consisting of an initial fitting phase and subsequent compression phase, (ii) The compression phase is relevant to the generalization of DNN, and (iii) compression phase occurs like a diffusion behavior. 
Using the previous DNN architecture on the modified NIST (MNIST) dataset, Saxe et al. \cite{saxe2019information} argued that the comprehensive nature of the compression phase depends on the choice of the activation function and that the behavior of the information plain is due to the nonlinearity of the activation function. 
%challenged the claim by Shwartz-Ziv and Tishby \cite{shwartz2017opening} that deep neural networks (DNNs) undergo a comprehensive compression phenomenon, arguing that this effect depends on the choice of the activation function. 
Recently, Noshad et al. \cite{noshad2019scalable} used a rate-optimal estimator of MI to show that an optimal hash-based estimator reveals compression across a broader range of networks, including those with rectified linear unit (ReLU) and max-pooling activations by training the fully-connected feed-forward network on the MNIST hand-written digits dataset. %On the other hand, Goldfeld et al. \cite{goldfeld2018estimating} argued that the observed compression results from geometric factors. However, 
Recent studies provided empirical \cite{geiger2021information} and theoretical \cite{kawaguchi2023does} evidence to support the above interpretations, using various training datasets and NN architectures with various hyper parameters. The IBT has since become an integral part of the analysis of deep learning architectures, particularly in optimizing cost functions \cite{alemi2016deep, voloshynovskiy2020variational} to achieve compressed representations.

The IBT was also used to analyze and design a deep AE \cite{yu2019understanding}, where the dynamics of stacked AE learning is bounded by specific data-processing inequalities, which are essential for the design of deep AE models. Three different training datasets were used with two different four-layered NNs that vary in the number of neurons per layer. %This will be covered in detail in section \ref{Sec:IBTDNN}. 
Taoia et al. \cite{tapia2020information} examined various optimization parameters and reported that an ideal AE with a large bottleneck layer size does not compress input information, whereas a small bottleneck size leads to effective compression in encoder layers. A new guideline was also proposed to adjust the parameters that compensate for scale and dimensionality effects.

Motivated by the above studies, the present work aims to understand how a non-stiff low-dimensional manifold (latent space), where the chemical kinetics evolve smoothly, is obtained using the AE-NODE architecture as reported in our previous study \cite{vijayarangan2023data}, thus serving as an effective surrogate model for the integration of stiff chemical kinetics. In particular, the present work attempts to answer the following questions: (i) What is the optimal dimension of latent variables? This requires an understanding of the relation between the intrinsic dimensionality of the physical problem and information compression; (ii) How latent variables are learned in dynamics-informed training? This requires an understanding of the gradient pathology of the network; and (iii) How are the physical variables transformed/combined to create the non-stiff latent manifold? This will be answered by understanding the disentanglement.  

The remainder of the paper is organized as follows. Section \ref{Sec:methodology} describes the NN architecture. Section \ref{Sec:Theory} discusses the key concepts of the IBT employed in the present study. The results are presented in Section \ref{Sec:ResultsandDiscussion} with a discussion of the important contributions of the present study. Finally, Section \ref{Sec:Conclusion} summarizes the work and suggests possible future work.

% ==============================================================================
% Sec:: Methodology
% ==============================================================================
\section{Neural network architecture} \label{Sec:methodology}

%The analysis methods employed in this work are presented in three sub-sections. In the first part, the neural network architecture (AE+NODE) and the loss function employed are briefed with the algorithm. In the second part, the elements of the information theory to the analysis of neural networks are explained. %and its application to the analysis and design of neural networks are explained. In the third part, the concepts of mode mixing and probability redistribution are detailed.

%This section describes the computational methodology employed in this study. First, the autoencoders and their working principle are briefed. Second, the concept of neural ODE used to integrate the ODEs using neural networks is discussed. Third, the neural network architecture used in this work, autoencoders combined with neural ODE, is explained.

%\subsection{Neural network architecture and training}\label{Sec:NN_architecture_training}

% ==============================================================================
% Sub:: AE
% ==============================================================================
\subsection{Autoencoders (AE)}\label{Sec:Autoencoders}
\begin{figure}[!ht]
    \centering
    \includegraphics[width=0.8\textwidth]{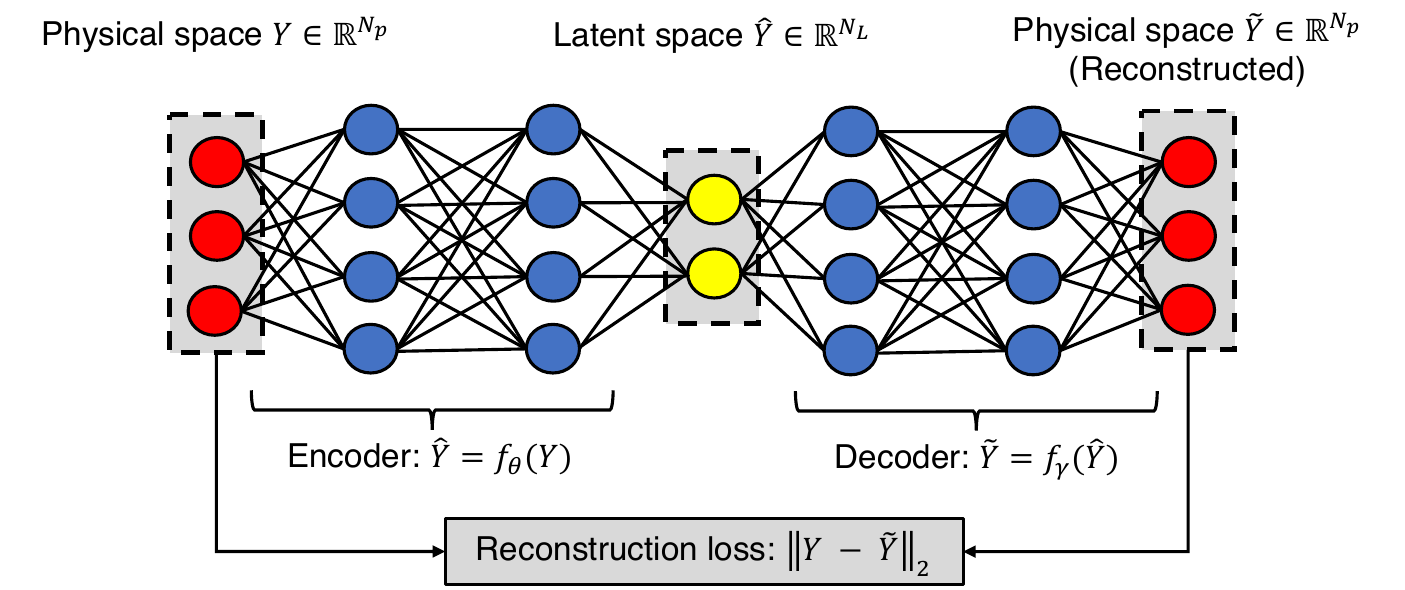}
    \caption{Schematic representation of a multi-layer autoencoder.}
    \label{fig:AE}
\end{figure}

An autoencoder (AE) is a feed-forward type network that transforms the variables from an input space to a reduced latent space with minimal distortion \cite{yu2019understanding}. Figure \ref{fig:AE} shows the schematic of an AE, which consists of two components: 1) a feedforward encoder that maps the physical space variables $( Y \in \mathbb{R}^{N_p} )$ to latent space variables $( \hat{Y} \in \mathbb{R}^{N_L} )$ using the mapping function $f_\theta$ and 2) a decoder which reconstructs to the original input space $( \tilde{Y} \in \mathbb{R}^{N_p} )$ from the latent space by using map ($f_\gamma$). Both mappings use a set of special operations comprised of a piece-wise linear function followed by a non-linear activation function. As such, the decoder is not an inverse operation of the encoder. 
% -------------------------------------------------------------
% Fig:: Autoencoders
%
% -------------------------------------------------------------

The encoder mapping function ($f_\theta$) is written as:
% Eq:: Encoder and Decoder
\begin{subequations} \label{eq:encoder}
\begin{align}
f_{\theta}(Y) &= g_{\theta_1} \circ g_{\theta_2} \circ \dots \circ g_{\theta_{N_h}}(Y) \\
g_{\theta_l}(T_l) &=  ELU \left( \theta^{w}_{l} \, T_{l-1} + \theta^{b}_{l} \right)\\
ELU(x) &= 
\begin{cases}
    x & \text{if $x > 0$} \\
    \alpha \times (\exp(x)-1) & \text{if $x \leq 0$}
\end{cases}
\end{align}
\end{subequations}
where $g_{\theta_l}$ represents the $l^{th}$ nonlinear piecewise operation. The parameters of the $l^{th}$ encoder layer, $\theta_l^w$ and $\theta_l^b$, denote the weights and biases, respectively, with $\theta_l^w \in \mathbb{R}^{N_l \times N_{l-1}}$ and $\theta_l^b \in \mathbb{R}^{N_l}$. The variable $T_l$ corresponds to the hidden variable in the layer $l$; specifically, $T_0 = Y$ when $l=0$, and $T_{N_h} = \hat{Y}$ when $l=N_h$. Among various options, the exponential linear unit (ELU) activation function, Equation \ref{eq:encoder}c, was employed with a coefficient $\alpha = 1.0$. The decoder mapping function ($f_\gamma$) is expressed in a similar way.
% -------------------------------------------------------------

%
% -------------------------------------------------------------
% -------------------------------------------------------------
This work focuses on a stacked autoencoder with a fully connected multilayer perceptron (MLP) architecture for both the encoder and decoder. Consider an input vector $Y \in \mathbb{R}^{N_p}$, where $N_p$ is the dimension of the solution variables. During training, the output of the AE $(\tilde{Y})$ is constrained to minimize the reconstruction error ($L_{AE} = \|Y - \tilde{Y}\|_2$) to match the input data $Y$. The middle layer, which represents the latent space and connects the encoder and decoder, is referred to as the bottleneck layer. The dimension of the bottleneck layer ($N_L$) is usually less than the dimension of the physical state ($N_p$) in the development of ROM, but can be greater than or equal to $N_p$~\cite{lee2021parameterized}.

%\times N_{sample} and $N_{sample}$ is the number of the data samples

A single-layer linear autoencoder, with the loss function $L_{AE}$, is essentially equivalent to the PCA process, although whether the operation will perform the least squares regression depends on the choice of loss functions. Interpretation of the general AE process by the linear transformation viewpoint is limited, and the deep nonlinear autoencoder theory is still an open research question. The nonlinear deep AE learns to project the data not onto a subspace, but onto a curvilinear manifold, which is the image of the encoder map or the preimage of the decoder map. The nonlinear activation for dimensionality reduction is the critical component to learn abstract features in an unsupervised manner, which can also be applied to a supervised task.

% ==============================================================================
% Sub:: Neural ODE
% ==============================================================================
\subsection{Neural ODE}
%This subsection briefly describes the working of neural network architecture Neural ODE, employed in this work. 
Neural ODE \cite{chen2018neural, rubanova2019latent} (NODE) is a deep learning framework  to approximate the continuous temporal dynamics of a system governed by the ODE of the form \footnote{For the present study, $Y$ and $y$ are used interchangeably.}
%
% -------------------------------------------------------------
% Eq:: ODE
\begin{equation} \label{eq:ODE}
\frac{dy(t)}{dt} = f(y(t), t) \,.
\end{equation}
%
% -------------------------------------------------------------
Employing any standard numerical method (e.g., the Euler method) to solve Eq. \ref{eq:ODE}, time integration is performed from an initial to the final time, during which $f$ is evaluated as a function of both $y$ and $t$. In the NODE framework, however, the hidden state $y(t)$ is the solution to the initial-value problem (IVP) and is governed by
%
% -------------------------------------------------------------
% Eq:: NODE
\begin{equation} \label{eq:NN}
\frac{dy(t)}{dt} = f_{\beta}(y(t), t)
\end{equation}
%
% -------------------------------------------------------------
where $f_{\beta}$ describes the dynamics of the hidden (temporal) state and is modeled using a NN with parameters $\beta$. Given an initial condition $y(t_0)$, Eq. \ref{eq:NN} can be integrated using any numerical ODE solver, as expressed in Eq. \ref{eq:NODE}, to obtain the predicted solution y$_\text{pred}$(t) at the desired time $t$ and to a required accuracy:
% -------------------------------------------------------------
% Eq:: NODE solution
\begin{equation} \label{eq:NODE}
y(t_1) = y(t_0) + \int_{t_0}^{t_1} f_{\beta} \left( y(t), t \right) dt = \text{ODESolve}\left( f_{\beta}, y(t_0), t_{0}, t_1 \right) \,,
\end{equation}
%
% -------------------------------------------------------------
% -------------------------------------------------------------
% Eq:: NODE Loss
\begin{equation} \label{eq:NODE_Loss}
L_\text{NODE} = \left|\left| y_\text{pred}(t) - y_\text{data}(t) \right|\right|_2 \,.
\end{equation}
%
% -------------------------------------------------------------
%$\left(\right)$
To minimize the scalar loss function (Eq. \ref{eq:NODE_Loss}) with respect to the NN parameters,  $\mathrm{d}L_\text{NODE}/\mathrm{d}\beta$, we need to identify how the loss depends on the hidden states $y(t)$ at each instant, $\mathrm{d}L_\text{NODE}/\mathrm{d}y(t)$. This quantity is  called the adjoint of the state, $a(t)$, whose dynamics are given by
% -------------------------------------------------------------
% Eq:: Adjoint
\begin{equation} \label{eq:NODE_Adjoint}
\frac{da(t)}{dt} = -a(t)^T \frac{\partial f_\beta (y(t),t)}{\partial y} \,,
\end{equation}
%
% -------------------------------------------------------------
% -------------------------------------------------------------
% Eq:: Sensitivity
\begin{equation} \label{eq:Sensitivity}
\frac{dL_\text{NODE}}{d\beta} = - \int_{t_1}^{t_0} a(t)^T \frac{\partial f_\beta (y(t),t)}{\partial \beta} dt \,.
\end{equation}
%
% -------------------------------------------------------------
Both of the vector-Jacobian products, $a(t)^T\frac{\partial f_\beta}{\partial y}$ and $a(t)^T\frac{\partial f_\beta}{\partial \beta}$ in Eq. \ref{eq:NODE_Adjoint} and \ref{eq:Sensitivity}, can be effectively evaluated by automatic differentiation. NODE employs an adjoint sensitivity method to formulate the augmented ODE system $\left[ \frac{dy(t)}{dt}, \frac{da(t)}{dt}, \frac{dL_\text{NODE}}{d\beta}\right]$. These augmented ODEs are integrated backward in time to compute the gradients. This approach results in a memory-efficient training of the neural network.
%
%
% ==============================================================================
% Subsec:: AE+Neural ODE 
% ==============================================================================
%
\subsection{Combined autoencoder and neural ODE architecture}\label{Sec:AE_NODE}
%
% -------------------------------------------------------------
% Fig:: AE&NODE
%
\begin{figure}[!ht]
    \centering
    \includegraphics[width=1.0\textwidth]{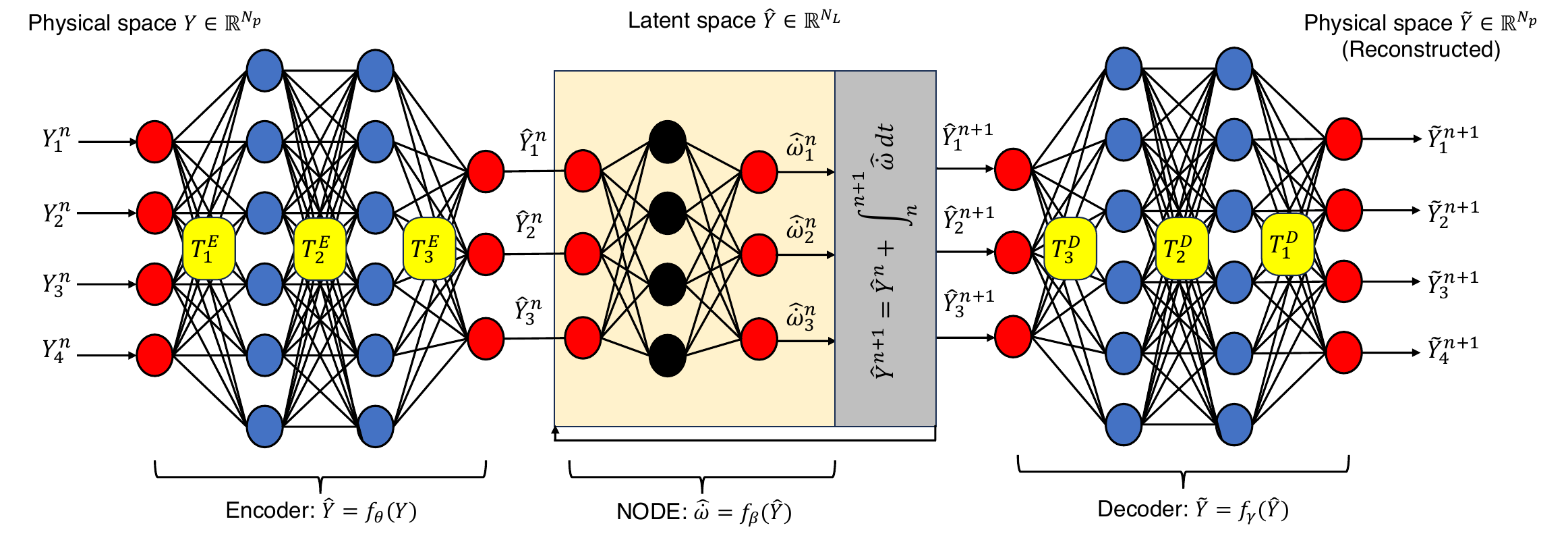}
    \caption{Schematic representation of autoencoder with neural ODE used in this work.}
    \label{fig:AENODE_arch}
\end{figure}
%
% -------------------------------------------------------------
%
In this work, we employ an AE integrated with the NODE as shown in Fig. \ref{fig:AENODE_arch}.  As explained in Sec. \ref{Sec:Autoencoders}, the solution variable vector at time $t_n$, $Y^n$, is mapped from a physical to a latent space, $\hat{Y}^n$, using the encoder. Then, the NODE advances the latent state variable from $\hat{Y}^n$ to $\hat{Y}^{n+1}$, using a numerical integration method. In this study, we employed the fourth-order explicit Runge--Kutta method. Finally, the decoder recovers the variables ($\tilde{Y}^{n+1}$) in the physical space. The dimension of the input and output vectors of the encoder was set to $N_{p} = 9$  and $N_L = 5$ (latent space dimension), respectively. Therefore, five latent variables were integrated utilizing the NODE, and the decoder retrieved them back to nine physical variables. 

The PyTorch library \cite{paszke2019pytorch} was used to train the neural network, which consisted of $N_h = 5$  hidden layers in the encoder, decoder, and NODE. Each hidden layer consisted of 100 neurons with the ELU activation to model the nonlinear chemical reaction rates. However, the output layer of the encoder, NODE, and decoder did not include any nonlinear activation units. Note that the latent variables $\hat{Y}$'s are obtained by the series of linear piecewise operation and nonlinear activation function of the variables ($~Y_{k}'s$), using the encoder. Therefore, there is no one-to-one correspondence between the variables ($~Y_{k}'s$) and ($\hat{Y}'s$). Thus, the training of the AE+NODE node requires special loss functions as discussed in Ref. \cite{vijayarangan2023data}, such as
% -------------------------------------------------------------
% Eq:: Loss -- AE + NODE
\begin{equation}\label{eq:Loss_AE_NODE}
\mathrm{Loss} = \varepsilon_1 L_1+ \varepsilon_2 L_2+ \varepsilon_3 L_3 \,.
\end{equation}

\begin{figure}[!ht]
    \centering
    \includegraphics[width=0.75\textwidth]{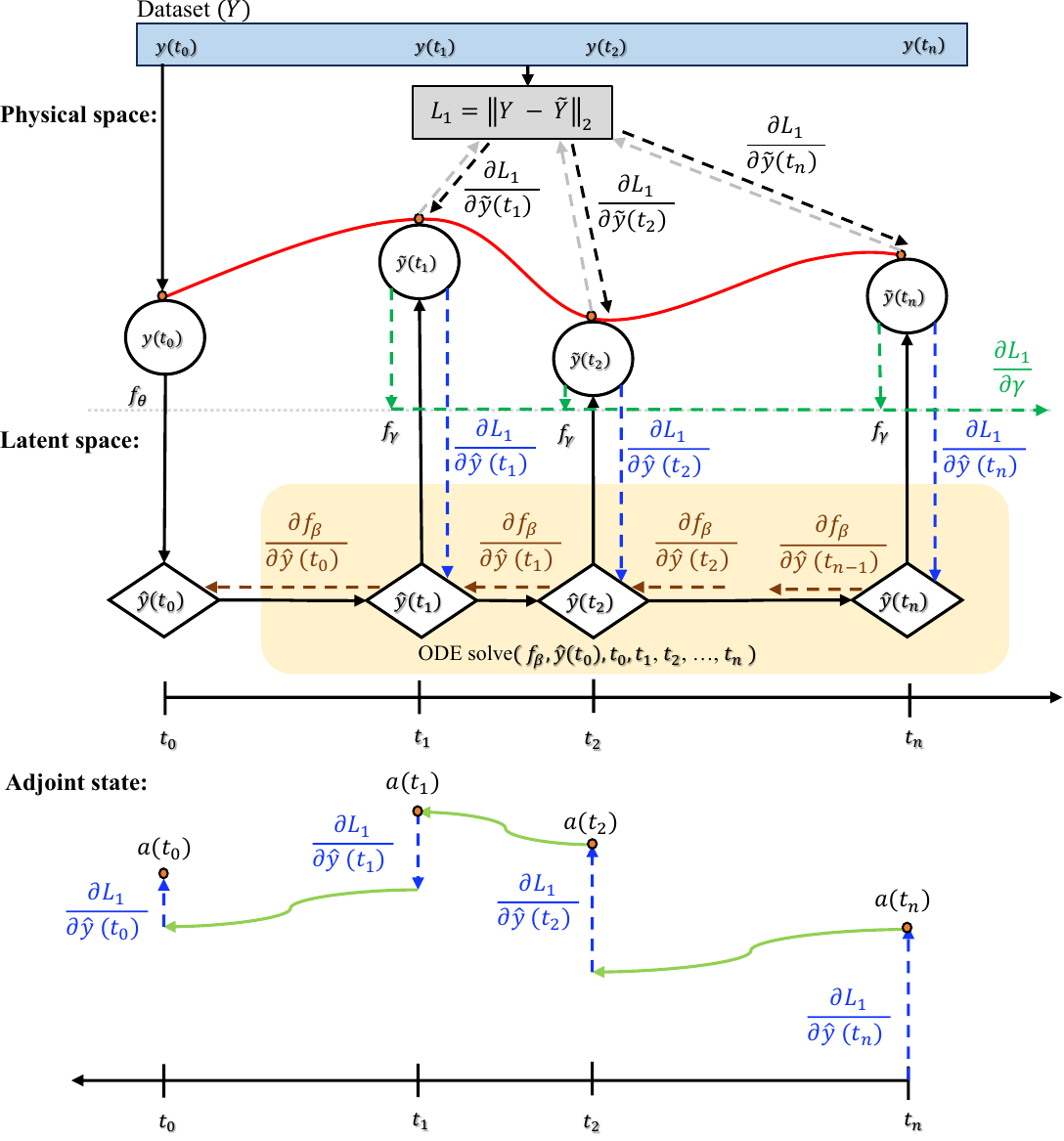}
    \caption{Schematic representation of the $L_1$ loss and the gradient flow for combined AE+NODE training.}
    \label{fig:L1_Loss}
\end{figure}

% -------------------------------------------------------------
% -------------------------------------------------------------
\begin{algorithm}[!ht]
    \caption{Reverse-mode derivative with $L_1$ loss function} \label{Alg1}
  \begin{algorithmic}[1]
    \INPUT AE+NODE parameters $\theta$, $\beta$, $\gamma$, start time (t$_0$), final time (t$_N$), solution state ($\tilde{Y}$)

    \STATE Compute $\frac{\partial L_1}{\partial \tilde{Y}}$, $\frac{\partial L_1}{\partial \hat{Y}}$, and $\frac{\partial L_1}{\partial \gamma}$ for the decoder part
    \STATE $\frac{\partial L_1}{\partial t_n} = \left( \frac{\partial L_1}{\partial \hat{y}(t_n)} \right)^T f_{\beta} \left(\hat{y}(t_n), t_n, \beta \right)$ 
        \Comment{Compute gradient w.r.t t$_n$}
    \STATE s$_0$ = $\left[\hat{y}(t_0), a(t), 0 , -\frac{\partial L_1}{\partial t_n} \right]$
        \Comment{Define initial augmented state}
    \STATE \textbf{def} \texttt{aug$\_$dynamics} $\left( \left[ \hat{y}(t), a(t), \_, \_ \right], t, \beta \right)$
        \STATE \textbf{return} $\left[ f_{\beta} \left( \hat{Y}, t, \beta \right), -a(t)^T \frac{\partial f_{\beta}}{\partial \hat{Y}}, -a(t)^T \frac{\partial f_{\beta}}{\partial \beta}, -a(t)^T \frac{\partial f_{\beta}}{\partial t}  \right]$
    \STATE $\left[ \hat{y}(t_0), \frac{\partial L_1}{\partial \hat{y}(t_0)}, \frac{\partial L_1}{\partial \beta}, \frac{\partial L_1}{\partial t_0} \right] = \texttt{ODESolve} \left( s_0, \texttt{aug$\_$dynamics}, t_1, t_0, \beta \right) $
    \STATE Compute $\frac{\partial L_1}{\partial y(t_0)}$ and $\frac{\partial L_1}{\partial \theta}$ using $\frac{\partial L_1}{\partial \hat{y}(t_0)}$ for the encoder part

    \OUTPUT $\frac{\partial L_1}{\partial \theta}$, $\frac{\partial L_1}{\partial \beta}$, $\frac{\partial L_1}{\partial \gamma}$, $\frac{\partial L_1}{\partial y(t_0)}$, $\frac{\partial L_1}{\partial t_0}$, and $\frac{\partial L_1}{\partial t_n}$
  \end{algorithmic}
\end{algorithm}
%

% -------------------------------------------------------------
\begin{algorithm}
    \caption{Reverse-mode derivative with $L_3$ loss function} \label{Alg2}
  \begin{algorithmic}[1]
    \INPUT Parameters $\theta$, $\beta$, start time (t$_0$), final time (t$_N$), final state ($\hat{y}(t_n)$), and loss gradient $\left( \frac{\partial L_3}{\partial \hat{y}(t_n)} \right)$
    
    \STATE $\frac{\partial L_3}{\partial t_n} = \left( \frac{\partial L_3}{\partial \hat{y}(t_n)} \right)^T f_{\beta} \left(\hat{y}(t_n), t_n, \beta \right)$ 
        \Comment{Compute gradient w.r.t t$_n$}
    \STATE s$_0$ = $\left[\hat{y}(t_0), a(t), 0 , -\frac{\partial L_3}{\partial t_n} \right]$
        \Comment{Define initial augmented state}
    \STATE \textbf{def} \texttt{aug$\_$dynamics} $\left( \left[ \hat{y}(t), a(t), \_, \_ \right], t, \beta \right)$
        \STATE \textbf{return} $\left[ f_{\beta} \left( \hat{Y}, t, \beta \right), -a(t)^T \frac{\partial f_{\beta}}{\partial \hat{Y}}, -a(t)^T \frac{\partial f_{\beta}}{\partial \beta}, -a(t)^T \frac{\partial f_{\beta}}{\partial t}  \right]$
    \STATE $\left[ \hat{y}(t_0), \frac{\partial L_3}{\partial \hat{y}(t_0)}, \frac{\partial L_3}{\partial \beta}, \frac{\partial L_3}{\partial t_0} \right] = \texttt{ODESolve} \left( s_0, \texttt{aug$\_$dynamics}, t_1, t_0, \beta \right) $
    \STATE Compute $\frac{\partial L_3}{\partial f_{\theta}(Y)}$ and $\frac{\partial L_3}{\partial \theta}$ for the encoder part
    
    \OUTPUT $\frac{\partial L_3}{\partial \theta}$, $\frac{\partial L_3}{\partial \beta}$, $\frac{\partial L_3}{\partial \hat{y}(t_0)}$, $\frac{\partial L_3}{\partial t_0}$, and $\frac{\partial L_3}{\partial t_n}$
  \end{algorithmic}
\end{algorithm}
\begin{figure}[!ht]
    \centering
    \includegraphics[width=0.75\textwidth]{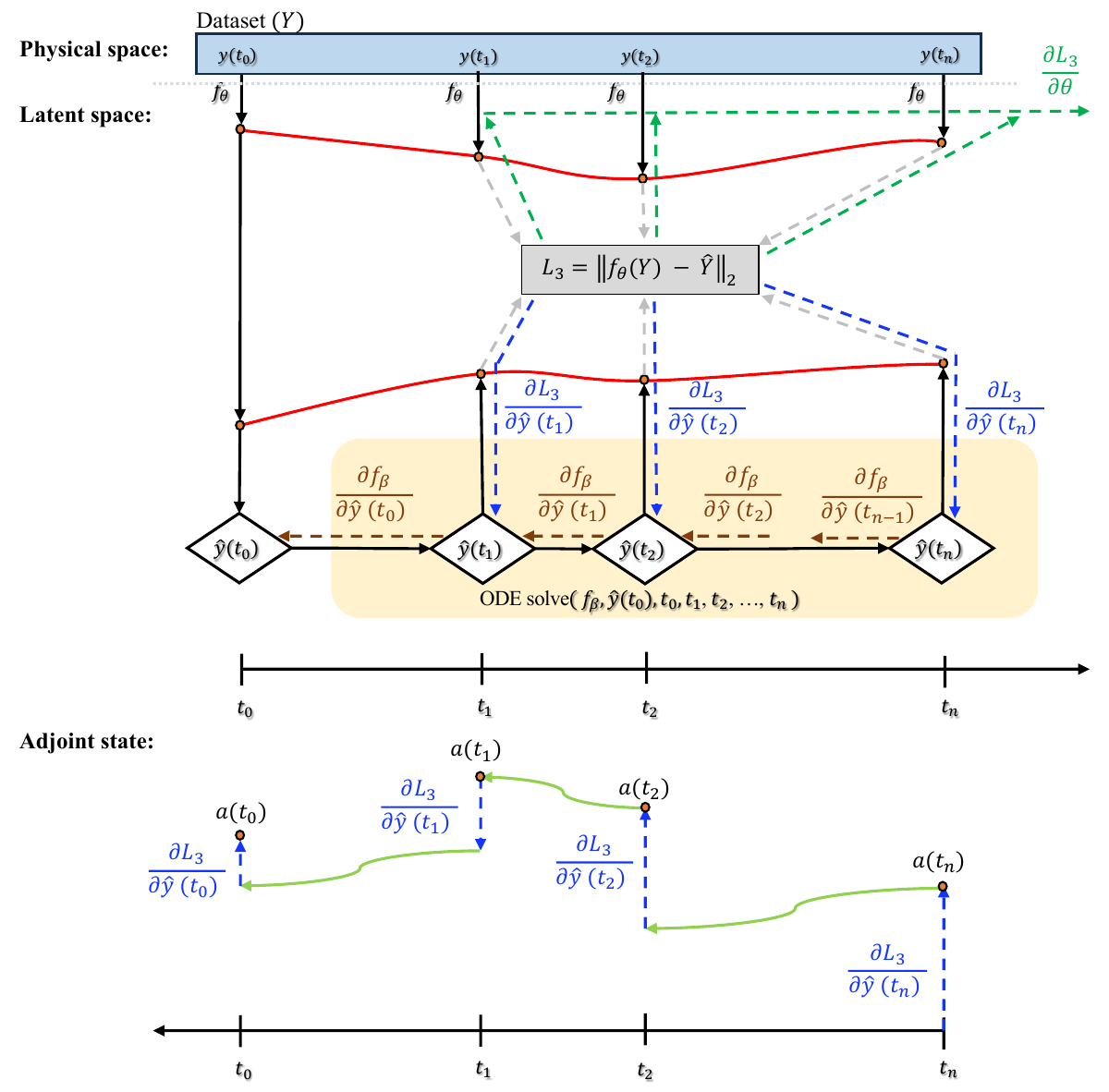}
    \caption{Schematic representation of the $L_3$ loss and the gradient flow for combined AE+NODE training.}
    \label{fig:L3_Loss}
\end{figure}

For the AE+NODE training, the reverse mode derivative computation for the given loss functions is crucial and dictates the property of the latent manifold. The training process is split into three steps where each step corresponds to a specific loss function in Eq. \ref{eq:Loss_AE_NODE}, as explained in the following. (i) $L_1 = \| Y - \tilde{Y} \|_2$ is the combined encoder+NODE+decoder loss term, and the computation of reverse mode derivative for the $L_1$ is described in Algorithm \ref{Alg1}, which is also illustrated in Fig. \ref{fig:L1_Loss}. (ii) $L_2 = \|f_\gamma(f_\theta(Y)) - \tilde{Y}\|_2$ is the encoder+decoder (AE) loss function without NODE, and the loss gradients with respect to encoder ($\theta$) and decoder ($\gamma$) parameters are computed using automatic differentiation through the Autograd library in PyTorch \cite{paszke2019pytorch}. (iii) $L_3 = \|f_\theta (Y)- \hat{Y}\|_2 $ is the encoder+NODE loss function without a decoder, which helps in ensuring the integrated NODE trajectory matches the encoder projected state. However, the reverse-mode derivative computation is different from the other two losses due to the loss function on the latent trajectory. The pictorial representation of the gradient computation is illustrated in Fig. \ref{fig:L3_Loss} and the algorithm is summarized in Algorithm \ref{Alg2}. Note that the additional loss terms $L_2$ and $L_3$ ensure the mapping from physical to latent space and vice versa is bijective (or one-to-one correspondence). Here, all the coefficients multiplying the loss terms are set to 1 ($\varepsilon_1 = \varepsilon_2 = \varepsilon_3 = 1$). A similar loss function has been incorporated in Ref. \cite{grassi2022reducing}. Once the gradients are computed, the parameters of the network are updated with the Adam optimizer \cite{kingma2014adam}. 
%

% ==============================================================================
% Theory
% ==============================================================================
\section{Information bottleneck theory and disentanglement}\label{Sec:Theory}

This section outlines the theoretical foundation of information bottleneck theory (IBT) principles and concepts of disentanglement that are applied to gaining understanding of neural network learning. The approach, initially proposed by Tishby et al. \cite{tishby2000information} and Saxe et al. \cite{saxe2019information}, leverages the concept of mutual information to gain insights into how neural networks learn. In this work, we extend this framework to explore the importance of latent dimensions in capturing physical representations and enhancing the learning of AE-based reduced-order models (ROM).

\subsection{Elements of information theory}\label{Sec:Elements_Info_theory}
%\subsubsection{Background}\label{Sec:Theory}
Statistical models are used for understanding complex data, thereby making predictions about unseen data. According to the statistical models, the estimation of $Y$ for a given $X$ involves extracting and using the information in $X$ relevant for the prediction of $Y$, which is equivalent to modeling $p(X,Y) = p(Y|X)p(X)$ with $p$ denoting the probability density function (PDF). Understanding $X$ requires more than just predicting $Y$, but it also requires identifying which features of $X$ play a role in the prediction \cite{zhang2020information}. The amount of relevant information in $X$ about $Y$ is quantified by the mutual information $I(Y;X)$. In the following, the general definition and the method to determine the mutual information are described based on the basic concepts of the information theory.

Consider two random variables, $X$ and $Y$. Let $X$ and $Y$ denote the information at the input and output layer of the neural network, respectively. The joint PDF of $X$ and $Y$ is denoted by $p(X,Y)$, and $p(X)$ and $p(Y)$ are their marginal probabilities, respectively. The mutual information, $I(X; Y)$, is a measure of interdependence between two random variables $X$ and $Y$ and defined as the relative entropy between the joint distribution and the product of marginal distribution~\cite{cover1999elements}, which is given by
%
%
%
% -------------------------------------------------------------
% Eq:: Mututal info
\begin{equation}\label{eq:MI}
    \begin{aligned}
        I(X;Y) &= H(X) - H(X|Y) = H(Y) - H(Y|X)\\
        I(X;Y) &= D_{KL}[p(X,Y) || p(X)p(Y)]
    \end{aligned}
\end{equation}
%
% -------------------------------------------------------------
%
%
% -------------------------------------------------------------
% Fig:: Mututal information
\begin{figure}[!ht]
    \centering
    \includegraphics[width=5.0cm]{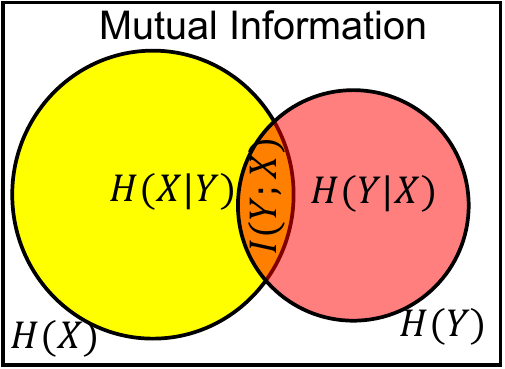}
    \caption{Schematic representation of mutual information from Shannon entropy.}
    \label{fig:IB}
\end{figure}
%
% -------------------------------------------------------------
%
%
where $H$ and $D_{KL}$ denote the Shannon entropy and Kullback--Leibler divergence (relative entropy), respectively.
Eq. \ref{eq:MI} implies that the mutual information reaches a minimum or zero when $X$ and $Y$ are independent and non-identical distributions. Figure \ref{fig:IB} shows the Venn diagram of the mutual information with the relation between entropy and conditional entropy. The intersection region represents the mutual information $I(X;Y)$. The relative entropy or Kullback--Leibler divergence is given by
%
%
% -------------------------------------------------------------
% Eq:: KL divergence
\begin{equation} \label{eq:KLDivergence}
D_{KL}[p(X) || q(X)]  = \int\int p(X) \log\left(\frac{p(X)}{q(X)}\right) dX
\end{equation}
%
% -------------------------------------------------------------
and the Shannon entropy ($H(X)$) is
% -------------------------------------------------------------
% Eq:: Shannon
\begin{equation} \label{eq:ShannonEntropy}
H(X) = - \int p(X) \log(p(X)) dX \,.
\end{equation}
%
% -------------------------------------------------------------
%
Therefore, the mutual information can be computed as
%
% -----------------------------------------------------------
% Eq:: Mutual Info.-1
\begin{equation} \label{eq:MutualInformation}
I(X;Y) = \int \int p(X,Y) \log\left( \frac{p(X,Y)}{p(X)p(Y)} \right) dX dY
\end{equation}
%
% -------------------------------------------------------------
%
%
which is the exact form to compute the mutual information. In practice, the evaluation of marginal and joint distributions depends on the choice of bin size which affects the outcome, $I(X;Y)$. Thus, an alternative to Eq. \ref{eq:MutualInformation} in determining the mutual information is sought using a non-parametric R\'enyi entropy formulation \cite{renyi1961measures}, written as:
% Eq:: Renyi entropy
\begin{equation} \label{eq:alpha_entropy}
H_{\alpha}\left(p\right) = \frac{1}{1-\alpha} \log \int_{x\in\mathcal{X}}p^{\alpha}\left(x\right) dx
\end{equation}
which is a generalized measure of the information while maintaining the additive nature of independent events. R\'enyi entropy is a generalization of Shannon entropy and the R\'enyi $\alpha$-entropy is defined for any continuous random variable $X$, with a probability density function $p(X)$ in a finite set $\mathcal{X}$. Here, the parameter $\alpha \in \mathbb{R}_{+}$ and $\alpha \rightarrow 1$ corresponds to Shannon entropy.
%
%
% -------------------------------------------------------------

% -------------------------------------------------------------
%
%Assuming $\alpha=2$ and approximating $\int_{x\in\mathcal{X}}f^{\alpha}\left(x\right)dx$ using the Gaussian Kernel density estimator leads to the quadratic Renyi entropy formulation given by:
%
%
% -------------------------------------------------------------
% Eq:: Quadratic Renyi
%\begin{equation} \label{eq:QuadraticRenyis}
%H_{2}(X) = -log\left( \frac{1}{N^2} \sum_{i=1}^{N}\sum_{j=1}^{N}G_{\sigma\sqrt{2}}\left(x_i - x_j\right)\right)
%\end{equation}
%
% -------------------------------------------------------------

Evaluating Eq. \ref{eq:alpha_entropy} is still a computationally intensive task. To overcome this limitation, a matrix-based estimate of R\'enyi entropy was proposed  \cite{giraldo2014measures} as a function that smoothly transforms the manifold of normalized positive definite (NPD) matrices to the actual numbers. Assuming a positive-definite kernel $\kappa: \mathcal{X} \times \mathcal{X} \rightarrow \mathbb{R}_{+}$ that is infinitely divisible, the Gram matrix, $K_{ij} = \kappa(x_i, x_j)$, is created by evaluating the kernel on all pairs of data points in $X = \{ x_1, x_2, ..., x_N \}$.
%($K$ an entropy-like function defined without estimating the probability density function of $X$) can be created by evaluating the kernel on all pairs of data points in $X = \{ x_1, x_2, ..., x_N \}$, where $K_{ij} = \kappa(x_i, x_j)$. 
Using the Gram matrix, an entropy-like function can be defined without the need to estimate the PDF of $X$.

%The study utilized the matrix-based Renyi's entropy formulation proposed in a prior study by Giraldo et al. \cite{giraldo2014measures}. The matrix-based estimator is a function that smoothly transforms the manifold of normalized positive definite matrices to actual numbers. By assuming a positively definite kernel $\kappa: \mathcal{X} \times \mathcal{X} \rightarrow \mathbb{R}_{+}$ that is infinitely divisible, the Gram matrix $K$ can be created by evaluating the kernel on all pairs of exemplars in $X = \{ x_1, x_2, ..., x_N \}$, where $K_{ij} = \kappa(x_i, x_j)$. Using the Gram matrix, an entropy-like function can be defined without the necessity of estimating the probability density function of $X$.

The matrix-based R\'enyi $\alpha$-entropy for an NPD matrix $A$ of size $N \times N$ and $tr(A)=1$ is written as:
%The function in Equation \ref{eq:MatrixRenyi} expresses the matrix-based Renyi's $\alpha$-entropy for a normalized positive definite (NPD) matrix $A$ of size $N \times N$, with $tr(A) = 1$. Where the NPD matrix $(A)$ is the normalized version of the Gram matrix K and $\lambda_i(A)$ denotes the i-th eigenvalue of A.
% -------------------------------------------------------------
% Eq:: Matrix Renyi
\begin{equation} \label{eq:MatrixRenyi}
H_{\alpha}\left(A\right) = \frac{1}{1-\alpha}\log_{2}\left(\sum_{i=1}^{N}\lambda_{i}\left(A\right)^{\alpha}\right)
\end{equation}
%
% -------------------------------------------------------------
where $\lambda_i$ represents the eigenvalues. In addition, the NDP matrix has the form
% Eq:: A matrix 
\begin{equation} \label{eq:AMatrix}
A_{ij} = \frac{1}{N}\frac{K_{ij}}{\sqrt{K_{ii}K_{jj}}} \,.
\end{equation}
%
% -------------------------------------------------------------
The matrix-based joint entropy of two random variables, $X$ and $Y$, is computed from Hadamard product $(A \circ B)_{ij} = A_{ij}B_{ij}$. Consider a sample $\{x_{i}, y_{i}\}_{i=1}^{n}$ from a pair of $n$ for two continuous random variables $X$ and $Y$, from the finite set $x\in\mathcal{X}$ and $y\in\mathcal{Y}$. The NPD matrices $A_{ij}$ and $B_{ij}$ are computed from Eq. \ref{eq:AMatrix} using the positive definite kernels $\kappa_1 : \mathcal{X}\times\mathcal{X} \rightarrow \mathbb{R}$ and $\kappa_2 : \mathcal{Y}\times\mathcal{Y} \rightarrow \mathbb{R}$, respectively. The product kernel $\kappa\left(\left(x_i,y_i\right),\left(x_j,y_j \right)\right) = \kappa_{1}\left(x_i,x_j\right)\kappa_{2}\left(y_i,y_j\right)$ is equivalent to $(A \circ B)_{ij}$. Then, the joint entropy is defined as
%
% -------------------------------------------------------------
% Eq:: Renyi joint matrix
\begin{equation} \label{eq:JointMatrixRenyi}
H_{\alpha}\left(A,B\right) = H_{\alpha}\left( \frac{A \circ B}{tr\left(A \circ B \right)}\right) \,.
\end{equation}
%
% -------------------------------------------------------------
Finally, the R\'enyi mutual information can be defined as
% -------------------------------------------------------------
% Eq:: Mutual info.--Renyi
\begin{equation} \label{eq:RenyisMI}
I_{\alpha}\left(X;Y\right) = H_{\alpha}\left( A\right) + H_{\alpha}\left( B\right) - H_{\alpha}\left( A,B\right) \,.
\end{equation}
%
% -------------------------------------------------------------
A key advantage of the applying matrix-based R\'enyi entropy formulation is its applicability to a more general dataset, such that $I_\alpha(X;Y)$ is computed directly from the NPD matrix, overcoming the issue of evaluating their corresponding densities. However, the free parameter kernel width, or window width, must be tuned by the smoothing parameter, $\sigma$, for proper estimation. In this work, $\sigma$ is defined by Silverman's rule \cite{silverman2018density}:
% -------------------------------------------------------------
% Eq:: Sigma
\begin{equation} \label{eq:sigma}
\sigma = h n^{-1/(4+d)}
\end{equation}
where $n$ is the sample size, $d$ is the dimensionality of the sample, and $h$ is the empirical constant based on the dataset. In this study, $h$ is tuned to match the dimensionality of the given data.

\subsection{Information bottleneck theory for DNN training}\label{Sec:IBTDNN}
% -------------------------------------------------------------
%Statistical techniques like mutual information and probability densities are to understand distribution learning and inference procedures. The basic concepts of mutual information and the method to find the metrics are already explored in section \ref{Sec:infoTh}. 

The basic concepts of the information theory discussed above are now applied to the DNN training process. 
The information $X$ (input) provides about the relevant quantity $Y$ (output) is squeezed through the bottleneck formed by the compressed (reduced-order) representation $\hat{X}$. As a result, the compressed representation $\hat{X}$ is used instead of $X$ in the prediction mapping problem. There must be a positive mutual information $I(X;Y)$. The information compression could lead to loss or no loss in fidelity while recovering the original data ($\tilde{X}$) from the compressed representation ($\hat{X}$). The compression process is called lossy if $ I(X;\tilde{X}) = H(X)$, or lossless if $ I(X;\tilde{X}) < H(X)$ (see Fig. \ref{fig:IB}). Therefore, $I(\hat{X};Y) \leq I(X;Y)$, since lossy compression cannot convey more information than the original data. While the goal of high fidelity ROM is lossless compression, a lossy compression (over-compression), $X \neq \tilde{X}$, is inevitable due to the incomplete data or lack of physical understanding. Moreover, the compressed representation, $\hat{X}$ of $X$, is not unique, and there can be multiple representations with an equal magnitude of mutual information (i.e., $I(X;\hat{X}_1) = I(X;\hat{X}_2) = \ldots$). However, only a few of those representations $\hat{X}_l$ may be interpretable. 

Compression is also interpreted as achieving a representation $\hat{X}$ of $X$ in terms of its minimal sufficient statistics (MSS). The selected representation $\hat{X}_l$ determines which information is preserved or lost. Thus, an optimal $\hat{X}$ should achieve an MSS of the joint distribution $p(X,Y)$ that relates the input $X$ to the outputs $Y$ (or the conditional distribution $p(Y|X)$ if $p(X)$ and $p(Y)$ are dependent).
Any representation $I(X;\hat{X}_i)$ can be transformed into an informationally-equivalent representation $I(X;\hat{X}_j)$ by invertible (non)linear transformations and rotations, also referred to as projection. With the goal of constructing a high-fidelity ROM that predicts the most relevant aspects of $Y$,  we need a sufficiently good, but parsimonious representation in the latent space.

% ==============================================================================
% Subsec:: Information Bottleneck Theory for AE
% ==============================================================================
\subsection{Information bottleneck theory for autoencoder}\label{Sec:IBTAE}
% -------------------------------------------------------------
In the context of the IB theory \cite{yu2019understanding, tishby2000information}, an AE compresses the input data and reduces the dimension, eventually reaching the information bottleneck. The encoder maps the physical variables ($Y$) as the input to latent space variables ($\hat{Y}$) using a nonlinear map $f_{\theta}(\hat{Y}|Y)$. The variables are retrieved back into the physical space ($\tilde{Y}$) from the latent space ($\hat{Y}$) by a decoder, which again employs a nonlinear map $f_{\gamma}(\tilde{Y}|\hat{Y})$ while minimizing the loss function $\|Y - \tilde{Y}\|_{2}$. This training process ensures that only relevant information from the incoming training data is retained while irrelevant information is discarded. 

Note that the AE process for ROM is specifically designed to reduce the dimensionality of the model in order to provide a \emph{compact} representation. 
%The data extraction process can be better understood using the IB theory, which shows that by limiting the capacity of the AE, one can constrain the model to extract only compact and useful representations in the training data. 
However, the definition of compact representation  depends on the context. Sometimes 
%Depending on the nature of the problem, reducing the dimension could represent the physics better; common ROMs use this concept, e.g., PCA. However, in a few cases, 
increasing the dimension may facilitate the analysis of the problem. For example, 
%That means increasing the dimension would help us understand the dynamics phase space clearly; this concept arrives from 
the Koopman theory transforms the input variables into the infinite-dimensional space in order to represent the complex nonlinear dynamics in a linearized form. 
%However, AEs are mainly utilized to reduce the dimension (which means the first kind mentioned above). 
While it has been demonstrated that AE successfully reproduces the temporal evolution of reacting systems by reduced-dimensional bottleneck latent space variables \cite{vijayarangan2023data}, there is no clear understanding of whether and how the latent variables retain the relevant information while removing fast time scales through the training process. To address such questions, the information flow in the AE framework is analyzed in the following.
%However, there is no clear study on using AE for dynamical systems other than their posterior prediction accuracy and the advantages of using lesser dimensions than the physical space. The present section explains the nature of the information flow in the AE, and the proceeding sections explain the natural way to explain the AE for dynamical systems.

For the feed-forward network, such as a stacked AE considered in this study, the layers of the encoder ($Y \rightarrow T^E_1 \rightarrow T^E_2 \rightarrow \dots \rightarrow T^E_5 \rightarrow T^E_6 = \hat{Y} $) and decoder ($\tilde{Y} \rightarrow T^D_1 \rightarrow T^D_2 \rightarrow \dots \rightarrow T^D_5 \rightarrow T^D_6 = \hat{Y} $) can be represented as a Markov chain \cite{yu2019understanding}. This implies that the forward and backward propagation are unidirectional, such that the variables are propagated from the input layer to the output layer, while the errors are back-propagated from the output layer to the input layer through the adjoint network. In other words, the output of the $\ell^{th}$ hidden layer depends only on the information present in the $(\ell-1)^{th}$ layer in the forward signal propagation, while the adjoint of the $(\ell-1)^{th}$ layer depends on the $\ell^{th}$ hidden layer in the backward propagation process. According to Yu and Principe \cite{yu2019understanding}, the AE network must satisfy the following two data processing inequalities (DPIs):
%
% Data processing inequalities (DPIs)
\begin{enumerate}
    \item Due to the symmetric architecture of the stacked AE
    \begin{enumerate}
        \item $I(Y;T^E_1) \geq I(Y;T^E_2) \geq \dots \geq I(Y;T^E_5) \geq I(Y;\hat{Y})$ \label{DPI1a}, in the encoder
        \item $I(T^D_1;\tilde{Y}) \geq I(T^D_2;\tilde{Y}) \geq \dots \geq I(T^D_5;\tilde{Y})\geq I(\hat{Y};\tilde{Y})$\label{DPI1b}, in the decoder
    \end{enumerate}
    \item The layer-wise mutual information must decrease with the depth of the network, \\ $I(Y;\tilde{Y}) \geq I(T^E_1; T^D_1) \geq \dots \geq I(T^E_6; T^D_6) = I(\hat{Y};\hat{Y}) =   H(\hat{Y})$ \label{DPI2}
\end{enumerate}
The DPI \ref{DPI1a} and \ref{DPI1b} imply that the hidden layers ($T^E_{\ell}$ and $T^D_{\ell}$) maximize the $\tilde{Y}$ (prediction) information while retaining the minimum relevant information in $Y$ (input/training data).

% ==============================================================================
% subSec: mode mixing and probability
% ==============================================================================
\subsection{Concept of disentanglement} \label{disentanglement}
Weierstrass's universal approximation theory asserts that a single-layer NN with infinite width can approximate any complex function. In practical applications, however, finite-width DNNs have been successfully employed to tackle complex problems. It has been empirically demonstrated that increasing the depth of a NN enhances its ability to approximate complex systems, as deeper layers lead to more robust representations. This phenomenon, known as deep representations, has been supported by many researchers. Bengio et al. \cite{bengio2013better} proposed an important hypothesis that disentangling the underlying factors of variation in data is crucial for improving the quality of learned representations in deep learning. According to this hypothesis, deep architectures are expected to be more effective when they isolate the distinct aspects of the data’s variability, leading to more interpretable and generalized models. The present subsection delves into this hypothesis, exploring how disentanglement plays a vital role in representation learning as outlined by Ref. \cite{bengio2013better}.

In practical applications, data often exhibit multiple modes that are separated by low-density regions. According to the manifold hypothesis, transitioning from these low-density regions (representing rare or unlikely events) to high-density regions (representing more probable events) poses a challenge in Markov processes. This difficulty highlights the importance of effective mixing between modes, especially in the deeper layers of a NN, where representation learning plays a critical role in capturing the underlying structure of the data. This mechanism is especially crucial in designing latent spaces for ROM development. The relationship between mode mixing and the depth of NNs has been hypothesized to play a pivotal role in learning disentangled representations. The mechanism behind this process can be understood through the following three hypotheses.

(i) Depth versus better mixing between modes: A successfully trained deep neural network architecture has the potential to generate representation spaces where the Markov chains mix more efficiently between distinct modes. As the depth of the network increases, it facilitates smoother transitions between low-density (rare event) and high-density (more probable event) regions, enabling better exploration of the underlying data structure.

(ii) Depth versus disentanglement: Deeper representations have a greater capacity to disentangle the underlying factors of variation in the data. Here, better disentangling implies that some of the learned features have a higher mutual information with some of the known factors. With increasing depth, neural networks can more effectively separate and isolate distinct features or patterns, leading to clearer and more interpretable latent spaces that capture the intrinsic variability within the data. 

(iii) Disentanglement unfolds and expands:  Disentangled representations not only (a) unfold the manifolds on which the data are concentrated, but also (b) expand the relative volume occupied by high-probability points near these manifolds. This expansion enhances the model’s ability to capture the most relevant aspects of the data, improving generalization and making the representation space more meaningful for downstream tasks.

The above theoretical basis will be employed in the subsequent analysis of the AE-NODE framework pertaining to our study.

% ==============================================================================
% Sec: Results
% ==============================================================================
\section{Results and discussion}\label{Sec:ResultsandDiscussion}
This section describes the information-theoretic view of AE for data-driven ROM of stiff dynamical systems. The main objective is to provide empirical evidence for how dynamics-informed training limits the architecture design of AE. This answers the question of how many latent space dimensions are required to model the stiff physical systems and how they influence the training process. The results are presented in five subsections: (i) Data acquisition and validation for dynamics-informed training; (ii) Information plane and data processing inequalities; (iii) Learning dynamics with dynamics-informed training; (iv) Disentanglement as a key to a smooth manifold; and (v) The effect of latent dimensions on training and prediction accuracy.

% ==============================================================================
% Subsec:: Data acquisition and validation for dynamics-informed training
% ==============================================================================
\subsection{Data acquisition and validation for dynamics-informed training}

As in our previous work \cite{vijayarangan2023data}, the model problem considered the H$_2$-air mixture with 10 species and 27 reactions \cite{mueller1999flow}, which is integrated by assuming a homogeneous constant-pressure batch reactor to obtain the ignition curve, using Cantera \cite{cantera}. As for the  thermodynamic conditions, pressure ($P$) is fixed at 1 atm, the initial temperature ($T_\mathrm{init}$) ranges from 1000 K to 2000 K, in steps of 100 K, and equivalence ratio ($\phi$) ranges from 0.5 to 1.5, in steps of 0.02.
The datasets from these conditions consisting of temperature and species mass fractions (except N$_2$ and Ar) are saved and randomly shuffled before the training. Furthermore, these datasets are split in the ratio 80:20 for training and testing, respectively, and normalized to accelerate the training process.

% -------------------------------------------------------------
% Fig:: SpeciesMassFraction&Temperature
\begin{figure}[!ht]
    \centering
    \includegraphics[width=0.9\textwidth]{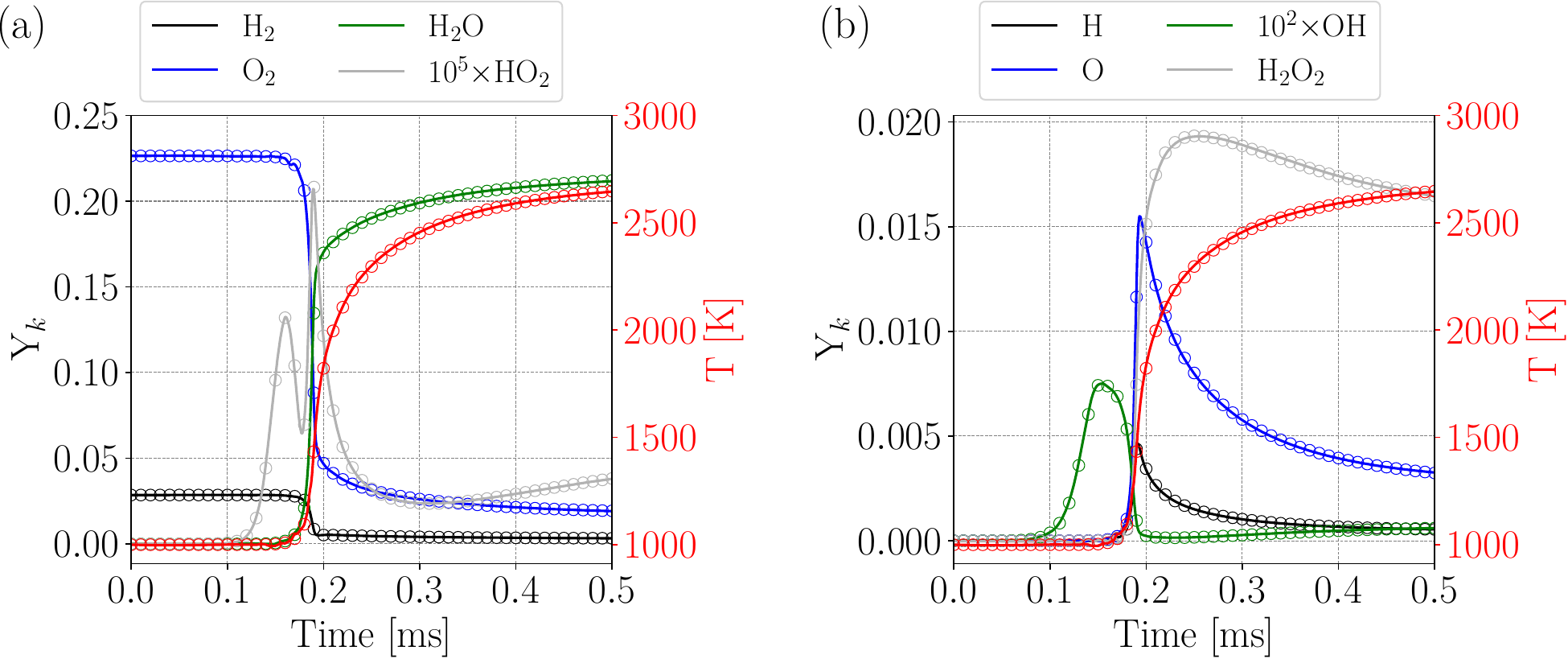}
    \caption{Comparison of temperature ($T$) and species mass fraction ($Y_k$) profiles for constant-pressure batch reactor ($P$=1 atm, $\phi$=1.0, $T_\mathrm{init}$=1000 K)  with H$_2$-air kinetics \cite{mueller1999flow} for Cantera (circles) vs. nonlinear AE+neural ODE (solid lines).}
    \label{fig:Yk_T_valid}
\end{figure}

% -------------------------------------------------------------
% Fig:: Latent space variables
\begin{figure}[!ht]
    \centering
    \includegraphics[width=0.45\textwidth]{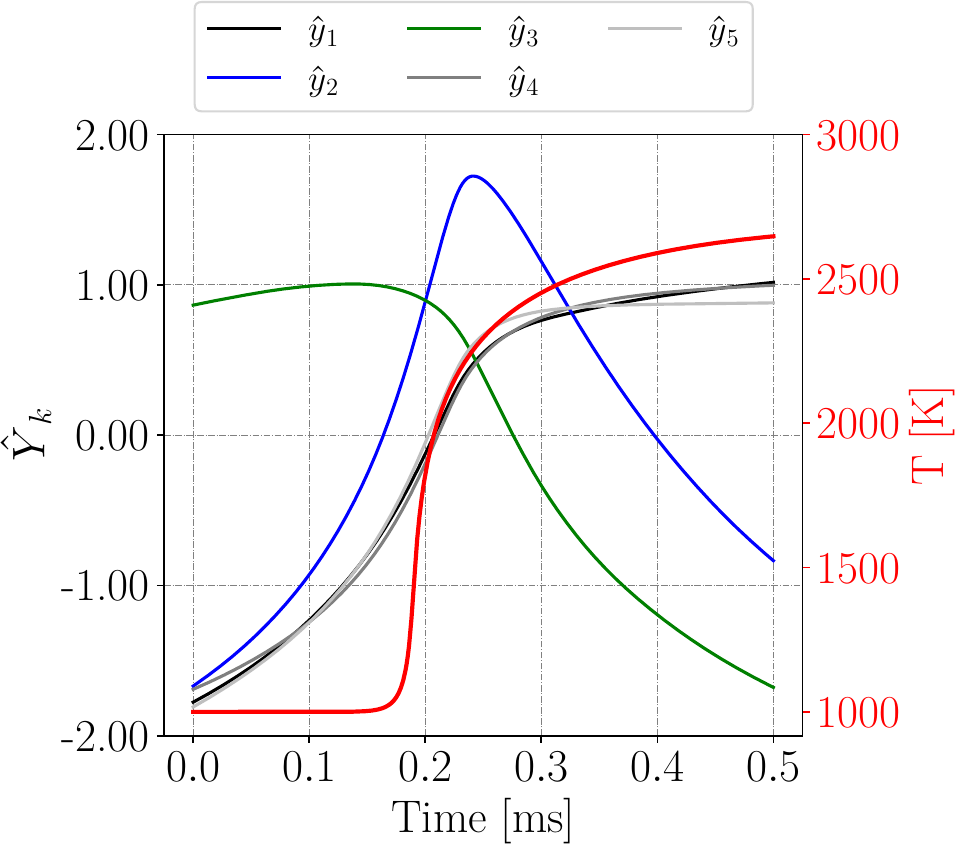}
    \caption{Time evolution of variables in the latent space for H$_2$-air kinetics reactor (at $P$=1 atm, $\phi$=1.0, $T_\mathrm{init}$=1000 K).}
    \label{fig:Latent_space}
\end{figure}

% -------------------------------------------------------------
As a first step, the present AE+NODE predictions were validated against the results obtained from direct integration with Cantera~\cite{cantera}. Figure \ref{fig:Yk_T_valid} compares the $T$ and $Y_k$ profiles obtained from the homogeneous batch reactor with the H$_2$-air mixture. For the $\phi$ = 1.0 and $T_\mathrm{init}$ = 1000 K values outside the training range, the predictions from the AE+NODE are in close agreement with the Cantera-based results for both major and minor species, demonstrating the prediction accuracy of the present approach. 

Fig. \ref{fig:Latent_space} shows the evolution of solution variables in the latent space. It is  evident that the latent variables evolve much more gradually in time without any sharp rise as seen in the major and minor species variables shown in Fig. \ref{fig:Yk_T_valid}. For N$_L$=5 latent variables  used for this study, the dynamics can be categorized into three distinct modes resembling  reactant ($\hat{y}_3$), product ($\hat{y}_1$, $\hat{y}_4$, $\hat{y}_5$), and intermediate species dynamics ($\hat{y}_2$). Similar behavior was observed for different chemical dynamics (e.g. C$_2$H$_4$-air) in Ref. \cite{vijayarangan2023data}. The following subsections systematically examine the impact of the latent dimension on dynamics-informed training in the design of AE-based ROMs from an information-theoretic perspective.
%
% ==============================================================================
% Subsec:: Information plane and data processing inequalities
% ==============================================================================
\subsection{Information plane and data processing inequalities}\label{Sec:IP_DPI}

In this subsection, an explanation for the learning dynamics of the AE is discussed using the concepts from information theory as described in section \ref{Sec:Theory}. The following discussion is presented for nonlinear AE(N$_L$=5)+NODE with dynamics-informed training. More general discussion for the AE design parameters is given in section \ref{Sec:IP_LD}

% -------------------------------------------------------------
% Fig:: Information Plane -- 1
\begin{figure}[!t]
    \centering
    \includegraphics[width=1.0\textwidth]{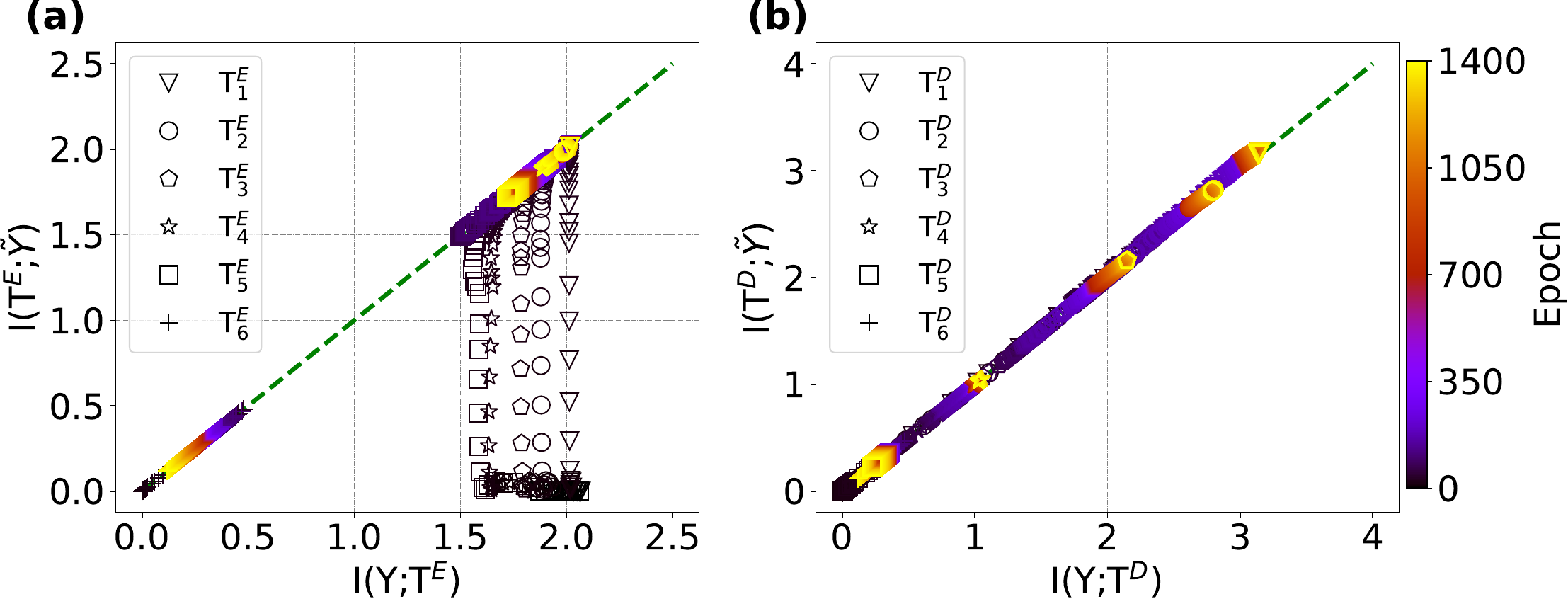}
    \caption{Information plane 1 (IP1) for the (a) encoder and (b) decoder, colored with the epoch. Symbols represent the different hidden layers (L = 5).}
    \label{fig:5Layer_L5}
\end{figure}
%Need to correct y axis label
% -------------------------------------------------------------

To verify the above two DPIs (\ref{DPI1a} and \ref{DPI1b}) and further understand the dynamics in the latent space, Fig. \ref{fig:5Layer_L5} shows the evolution of the correlations between the input and output mutual information (MI) with epochs at each hidden layer for the (a) encoder and (b) decoder, respectively. An epoch is counted when the current testing loss is lesser than its previous value. Both Figs. \ref{fig:5Layer_L5}(a) and (b) compare the amount of information that each $\ell^{th}$ hidden layer preserves about the input $\left(I(Y;T^E_{\ell})\right.$ and $\left.I(Y;T^D_{\ell})\right)$ with respect to the predicted output $\left(I(T^E_{\ell};\tilde{Y})\right.$ and $\left.I(T^D_{\ell};\tilde{Y})\right)$. 

%Fig. \ref{fig:5Layer_L5}a shows that at an early stage of training, also called the fitting phase, the hidden layer ($T^E$) contains much more information from the input ($Y$) than the output ($\tilde{Y}$) could obtain from the hidden layer, i.e. $I(Y;T^E_{\ell}) \gg I(Y;T^E_{\ell})$. The situation improves over the epochs, reaching the 45 degree bisector line (green dashed line), implying that predicted output retrieves nearly the same level of information. %It is evident that the first layer ($\ell=1$) has the farthest gap in the amount of the information and thus shows the largest distance from the bisector line. Once the training has reached the final state, all MI values reach the bisector line. At the end of the training (shown as yellow symbols), a compression in the MI is seen with the depth in the hidden layer from 1 to 5, with its value decreasing along the bisector line. A significant level of compression is seen at the final layer ($\ell=5$). 

Fig. \ref{fig:5Layer_L5}a shows two distinct phases that occur during the training of the neural network. The first and short phase is called the fitting phase, and the second and relatively longer phase is called the compression phase. The fitting phase is characterized by a sharp increase in $I(T^E_{\ell};\tilde{Y})$. As this curve reaches the bisector line (dashed green line),  both $I(Y;T^E_{\ell})$ and $I(T^E_{\ell};\tilde{Y})$ decrease along the bisector line as the redundant data are discarded. This stage is called the compression phase, where the local representations are fine-tuned. These results corroborate the observations made in Ref. \cite{yu2019understanding, tishby2015deep}. In a well-trained AE, the final value of $I(T^E;\tilde{Y})$ for each layer tends to be close to the value of $I(Y;T^E)$. In other words, the final points $(I(Y;T^E), I(T^E;\tilde{Y}))$ on the information plane should touch the limiting line for better prediction dictated by the fitting phase. In addition, the compression after the completion of the  fitting phase is responsible for fine-tuning the accurate prediction.

During the initial epochs, the hidden layer close to the input has higher $I(Y;T^E_{1})$ compared to the subsequent layers and negligible $I(T^E_{1};\tilde{Y})$. This implies that the first hidden layer has maximum information about the input data and negligible information about the output. Although $I(T^E;\tilde{Y})$ starts near zero during initial epochs for all hidden layers, it increases sharply with training, and the rate of increase drops with the depth of the network. This suggests that the layer close to the input (outer layers) learns faster than the subsequent/deeper layers. DPI \ref{DPI1a} further informs that the entropy of the information increases with the depth of the encoder, which implies that, as the input data cascade through the hidden layers, only a part of the information is retained and the remaining part is discarded. Therefore, as the training set consisting of $Y_k \in \mathbb{R}^{N_p}$ ($N_p = $ 9) cascades through the encoder and gets compressed to five latent variables, the compression of only the relevant information required to predict the dynamics (e.g., ignition curve) is retained, while the rest is filtered out. In other words, the information entropy of latent variables is always less than that of the input variable. From the basic understanding of AE, one can expect a near-equal information gap between consecutive hidden layers. In contrast to Ref. \cite{yu2019understanding}, for the dynamics-informed training, the information gap between hidden variables is not near-equal. Moreover, the compression phase does not occur in a single step; it is because the network adjusts the complexity of hidden variables to improve accuracy.

Fig. \ref{fig:5Layer_L5}b shows the mutual information curves for the decoder \footnote{The magnitude of the MI values vary (no normalization has been done) in both the encoder and decoder IP-1 due to a common choice of smoothing parameter, Eq.~\ref{eq:sigma} ($\sigma$).}, starting near zero due to random initialization. During the fitting phase, the MI points move along the bisector line as the decoder learns both the input $Y$ and the output $\tilde{Y}$. After the fitting phase (during the compression phase), it is desired to have $\tilde{Y}$ closer to $Y$. During the compression phase, the MI $ \left( I(Y;T^D_{\ell}), I(T^D_{\ell};\tilde{Y}) \right) $ of the inner layers drops due to the compression. However, the outer layers perform minor weight corrections in the compression phase to accurately predict the output. Therefore, small variations in the information plane are observed. In contrast to the encoder, the information gap between the hidden layers is well established; in other words, the hidden layers of the decoder contribute nearly equal work to learning the projection operation. The reason for the encoder's projection workload not being equally balanced between their hidden layers will be explained in subsections \ref{Sec: Learning_Dynamics} and \ref{Sec:IP_Disentanglement}

%The level of discrepancies in the earlier stage of fitting is not as much as that in the encoder training. 
%After the fitting phase is completed, the MI point for each hidden layer moves up along the bisector line (from 5 to 1), implying that the information is \emph{decompressed} as the training progresses to yield more accurate retrieval of the output. 
%which is ensured by minor weight corrections in the outer layers of the decoder. 
%During the compression phase, MI $ \left( I(Y;T^D_{\ell}), I(T^D_{\ell};\tilde{Y}) \right) $ of the inner layers decreases due to the compression occurring in the encoder. 
% -------------------------------------------------------------
% Fig:: Information Plane -- 2
\begin{figure}[!ht]
    \centering
    \includegraphics[width=0.6\textwidth]{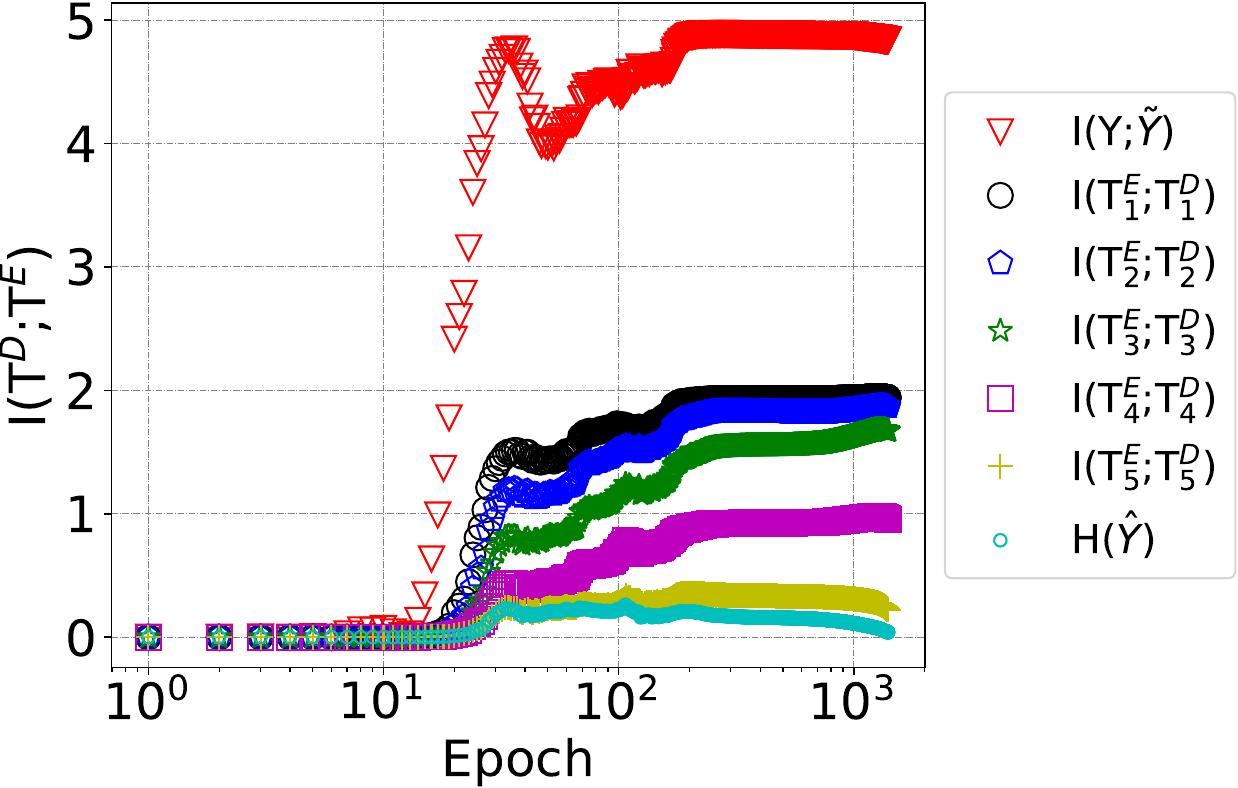}
    \caption{Information plane 2 (IP-2): Evolution of layer-wise mutual information with epoch.}
    \label{fig:5Layer_L5_plnae2}
\end{figure}
% use {Figure/InformationPlane_log.pdf} for log y axis
% -------------------------------------------------------------

%Analyzing the MI presented using IP-1 partially describes how the complexity changes during encoding. Therefore, 
To provide further insights into the process of complexity reduction, the evolution of MI during the training process for each hidden layer pair, $I(T^E_{\ell};T^D_{\ell})$, is plotted in Fig. \ref{fig:5Layer_L5_plnae2}, referred to as the information plane 2 (IP-2). 
%A properly designed AE should satisfy the DPIs \ref{DPI1a}, \ref{DPI1b}, and \ref{DPI2}. The layer-wise MI of the hidden layers ($I(T^E_{\ell};T^D_{\ell})$) 
It is seen that inequality DPI \ref{DPI2} is clearly satisfied, and MI decreases with the depth of the AE. Moreover, it describes the complexity of the hidden variables and also shows that they are bounded between the outer layer MI ($I(Y;\tilde{Y})$) and the latent space MI ($I(\hat{Y};\hat{Y})$) as given by DPI \ref{DPI2}. 
%Fig. \ref{fig:5Layer_L5_plnae2} shows that the layer-wise mutual information 
The linear activation used in the bottleneck layer implies $I(\hat{Y};\hat{Y})=H(\hat{Y})$, and $I(Y;\tilde{Y}) \rightarrow H(Y)$ as MSE loss$\rightarrow 0$. During the fitting phase, with an increase in the epochs, the MI value increases after some initial fluctuations. 
%As in the previous IP-1 map, at the final stage of the training the level of information compression is clearly seen as the depth of the layers increases from 1 to 5.
After the fitting phase, the layer-wise MI for the outer and inner layers shows different trends. The layer-wise MI for the outer layers flattens, as further gain in information is not possible once the predictions match with the ground truth. Additionally, during the compression phase, the variables in the innermost layers undergo transformation such that the layer-wise MI of these layers decreases. The existence of the compression phase depends upon the network's ability to ensure lossless reconstruction. In summary, IP-2 and DPI \ref{DPI2} state that the deeper the AE, the more the information is lost in the hidden layers, which leads to the loss of local structure of the input data and thereby changes the distribution of the hidden/latent representation. These results are in line with the observations made in Ref. \cite{yu2019understanding}.

From the above discussion, it is evident that deeper representations reduce the complexity of the input data, resulting in smoother dynamics in the latent space by changing their distributions. Fig. \ref{fig:Latent_space} validates this argument by showing smooth dynamics for large hidden layers and non-smooth for smaller hidden layers, respectively, as reported in Ref. \cite{vijayarangan2023data}. Thus, for a nonlinear AE+NODE with N$_L=5$, the complex evolution of species and temperature profiles in the physical space (Fig. \ref{fig:Yk_T_valid}) are transformed to smooth representations in the latent space (Fig. \ref{fig:Latent_space}).

% ==============================================================================
%Sub-sec:: Learning dynamics with dynamics-informed training
% ==============================================================================
\subsection{Learning dynamics with dynamics-informed training}
\label{Sec: Learning_Dynamics}

% -------------------------------------------------------------
% Fig:: Hidden variables 
\begin{figure}[!t]
    \centering
    \includegraphics[width=1.0\textwidth]{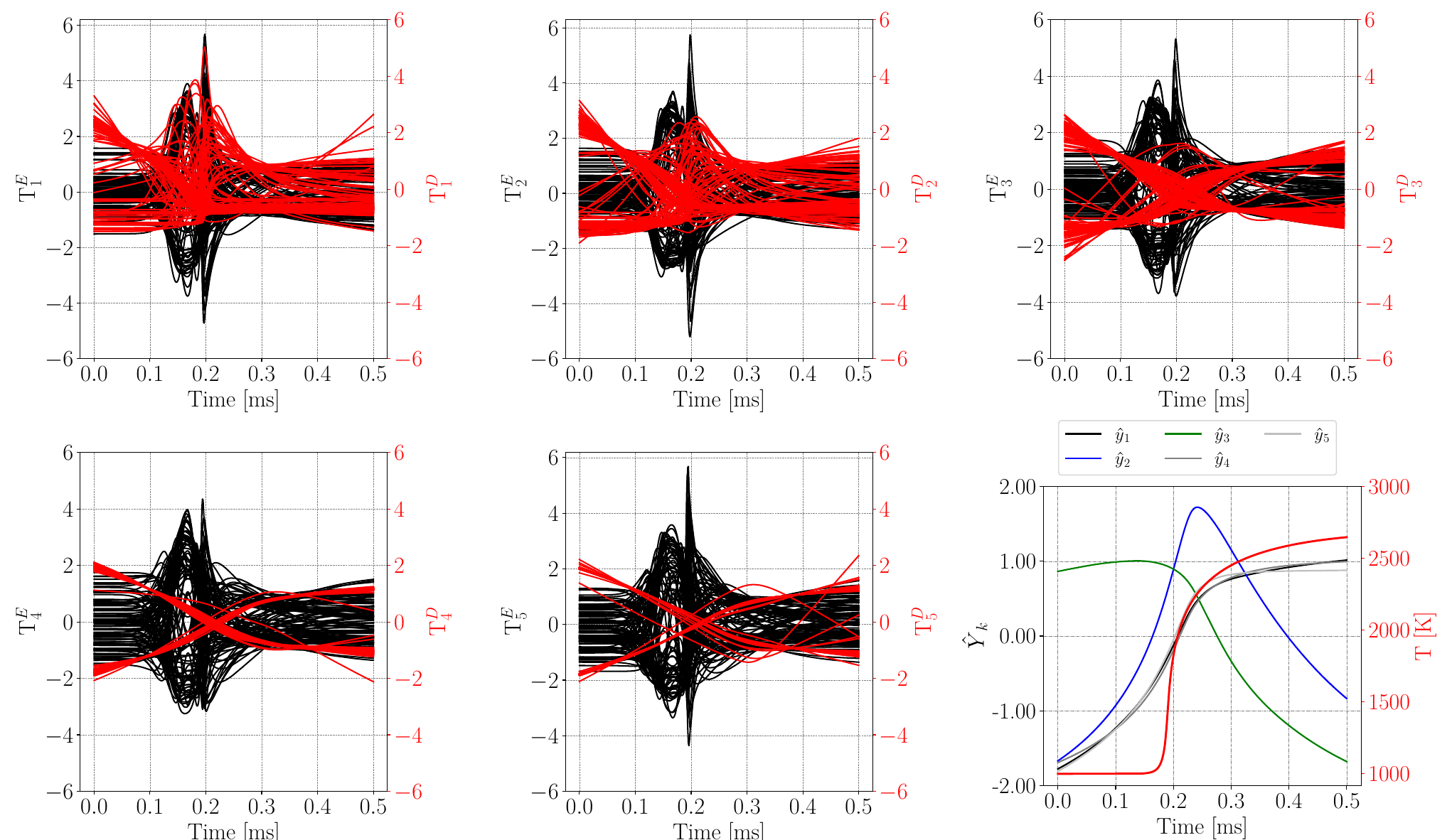}
    \caption{Comparision of the symmetric layerwise hidden variables dynamics for both encoder (left y-axis and colored in black) and decoder (right y-axis and colored in red).}
    \label{fig:Hidden_variables}
\end{figure}
% -------------------------------------------------------------

The previous subsection demonstrated the compression of the information through the AE+NODE. However, it remains unclear how the latent space dynamics are learned during the combined AE+NODE training. The key question is to understand whether the encoder filters unnecessary information or the decoder primarily contributes during the training phase,  and why the mappings of the encoder and decoder are not symmetric.

To investigate this, the temporal evolution of all hidden variables in each layer of the encoder (black) and decoder (red) is shown in Fig. \ref{fig:Hidden_variables} for the fully trained network. It is clear that the encoder and decoder mappings are not symmetric in dynamics-informed learning. This asymmetry arises due to the loss function in Eq. \ref{eq:Loss_AE_NODE}, which does not enforce layer-by-layer symmetric operations. Among the various loss terms, the $L_2$ term constrains the AE to predict the output as close as possible to the input. However, it does not necessarily reduce the stiffness in the latent space. In contrast, the $L_3$ term ensures that the encoder-mapped latent variables yield one-to-one correspondence with the neural ODE trajectory in the latent space or vice versa. Therefore, it is noted that the primary contribution to learning the smooth latent manifold comes from the $L_1$ term. 
%The study investigates how these latent trajectories are made less stiff compared to the physical state.

During the initial stages of training, the encoder maps the initial values from the physical space to the latent space while performing forward propagation, and NODE integrates the trajectory for the given final time. At this point, the only information NODE knows about the physical space is the initial condition, which leads to random initial maps. The entire trajectory is then mapped back to the physical space using the decoder, and the reconstruction loss $L_1$ is computed. The gradient of the loss function with respect to the network parameters is computed as described in Algorithm \ref{Alg1} and illustrated in Fig. \ref{fig:L1_Loss}. By this stage, the gradient of the last decoder layer contains more information than the subsequent layers (as discussed in subsection \ref{Sec:IBTDNN}), due to the unidirectional flow of information in both forward \cite{yu2019understanding} and backward passes \cite{chang2022explaining}. Consequently, the level of information modeled by T$_D^1$ is higher than the one by T$_D^2$, and that of T$_D^2$ is higher than the one by T$_D^3$, etc. Therefore, the latent trajectory is learned through the back-propagation of the decoder rather than the forward propagation of the encoder. This is evident from Fig. \ref{fig:Hidden_variables}, where the stiffness of the trajectory increases from the decoder's input ($\hat{Y}$) to its output ($\tilde{Y}$). However, there is an abrupt jump in the complexity of the dynamics in the hidden layer T$^E_5$ and $\hat{Y}$. 

Although the present subsection pointed out the corresponding gradient pathology for the smooth manifold learning process, it is unclear how the encoder projection is abrupt in the $\ell$=5$^{th}$ layer, and the decoder projections remain gradual. The next subsection will address this question by examining the disentanglement mechanism of the neural network.

%Subsection \ref{Sec:IBTDNN} illustrated the training dynamics of ignition problems using the information plane. %to understand how the encoder filters out temporal stiffness during the fitting and compression process. 
%This subsection presented the rationale behind the reduced stiffness in the latent space. The next subsection will discuss why this reduction occurs by examining the disentanglement mechanism of the neural network.

% ==============================================================================
%Sub-sec:: Disentanglement as a key to a smooth manifold
% ==============================================================================
\subsection{Disentanglement as a key to a smooth manifold}\label{Sec:IP_Disentanglement}

% -------------------------------------------------------------
% Fig:: PDF -- Physical variables
\begin{figure}[!ht]
    \centering
    \includegraphics[width=0.75\textwidth]{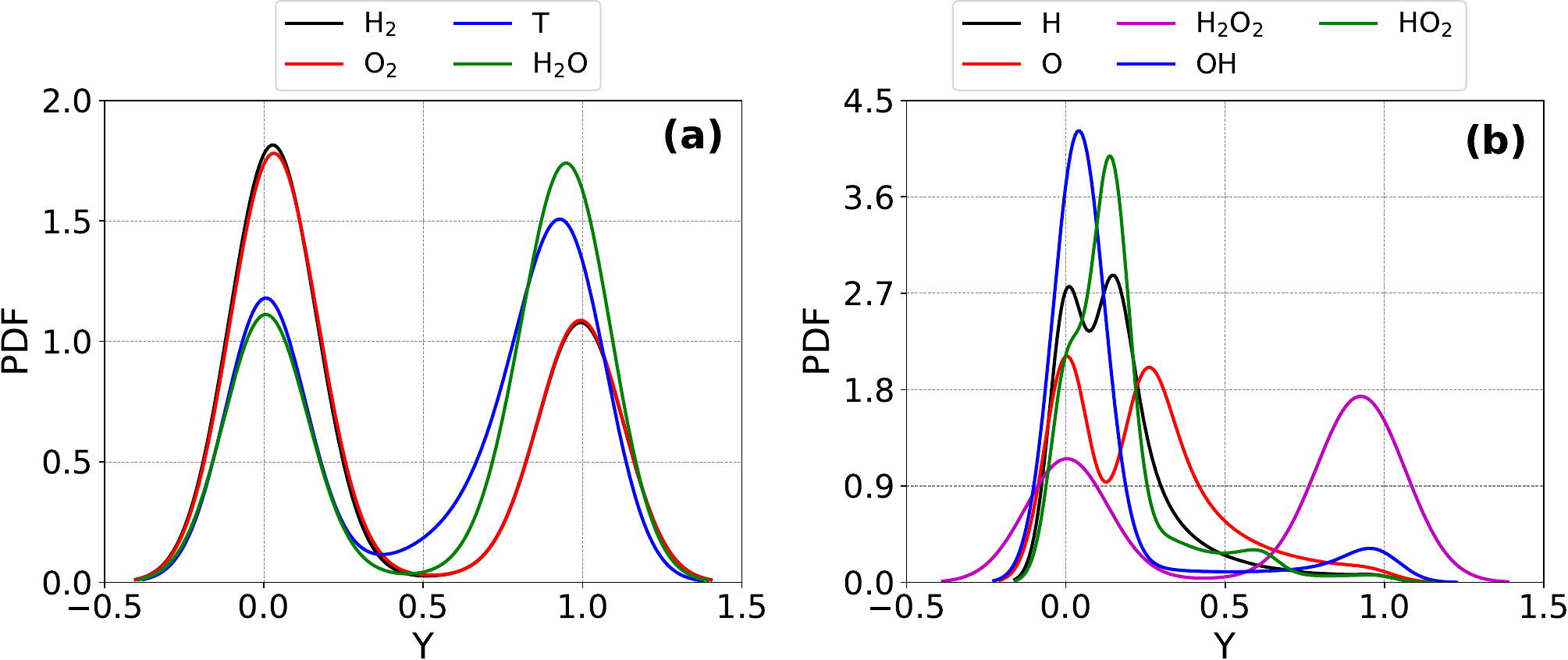}
    \caption{Probability density function (PDF) of the normalized (a) major species and temperature, and (b) minor species, during the ignition of H$_2$–air mixture at $P=1$ atm, $T_\mathrm{init}$=1000 K, and $\phi$=1.0.}
    \label{fig:PDF_phyVars}
\end{figure}

% -------------------------------------------------------------
% Fig:: PDF -- Latent variables
\begin{figure}[!ht]
    \centering
    \includegraphics[width=0.4\textwidth]{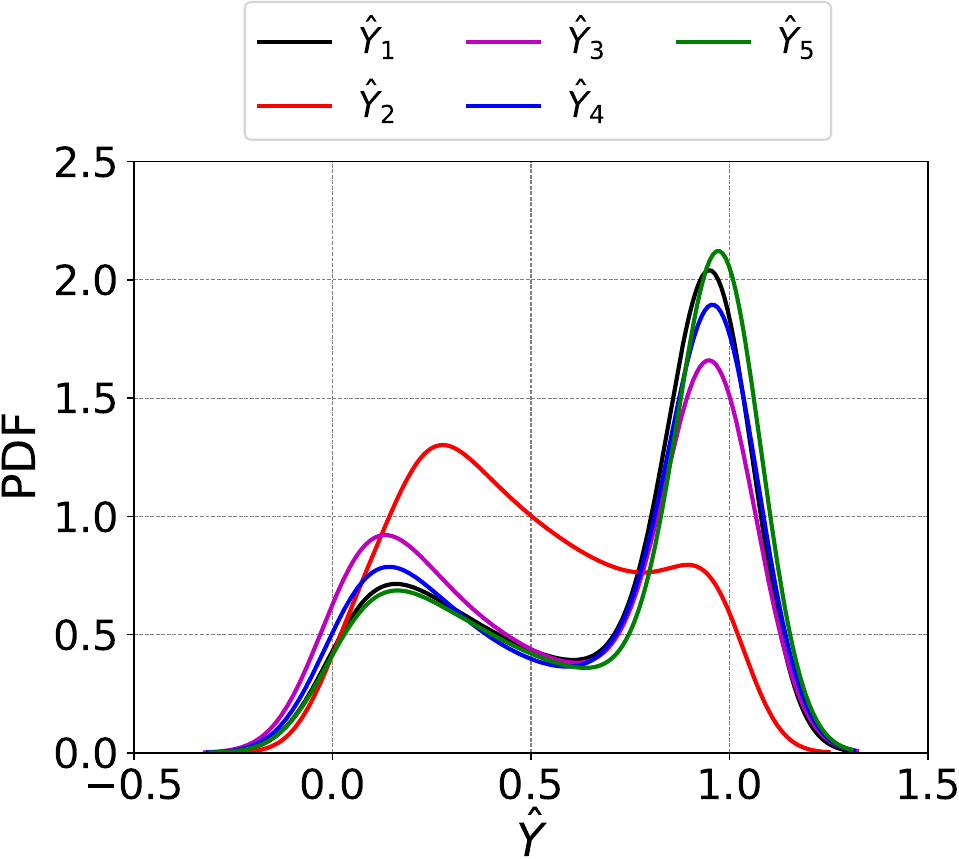}
    \caption{Probability density function (PDF) of the normalized latent variables during the ignition of H$_2$–air mixture at $P=1$ atm, $T_\mathrm{init}$=1000 K, and $\phi$=1.0 for nonlinear AE ($L=5$)+NODE.}
    \label{fig:PDF_LatVars}
\end{figure}
% -------------------------------------------------------------

Fig. \ref{fig:PDF_phyVars} and \ref{fig:PDF_LatVars} compare the density distribution (from kernel density estimation) of the variables in physical and latent spaces, respectively, during the ignition of H$_2$-air mixture. In the physical space (Fig. \ref{fig:PDF_phyVars}a and b), the PDF shows nearly a bi-modal distribution with two dominant peaks corresponding to unburnt and burnt equilibrium regions (high probable events) separated by ignition as a less probable event. Upon considering the AE+NODE into the bottleneck layer using the nonlinear activation, Fig. \ref{fig:PDF_LatVars} shows that the PDF is much smoother, implying that the ignition is transformed into a more probable event. This is attributed to the reduced complexity in the latent space, as explained by the IB theory. To explain how these distributions are formed and the low-probable  event becomes a more probable one, we refer to the better-mixing perspective of Bengio et al. \cite{bengio2013better}. In particular, the main focus is given to the reason behind the abrupt and gradual changes in the mapping of the encoder and decoder hidden layers.

The neural network architecture, such as the one considered in this study, forms a Markov chain. It is well understood that the Markov chain (shallow neural networks) faces challenges in jumping from one mode to another when separated by a large low-density region representing a rare or low-probable event, such as ignition. However, choosing a network of sufficient depth will result in a representation space that can disentangle the underlying factors of variation and a smoother evolution of dynamics in the latent space (see Fig. \ref{fig:Latent_space}). This overcomes the problem of mixing between modes by i) expanding the volume occupied by high-probability points (unburnt and burnt regions) and ii) unfolding the manifold where the raw input data (unburnt and burnt regions) concentrate as shown in Fig. \ref{fig:PDF_LatVars} and described in Ref. \cite{bengio2013better}. 

In brief, the dynamics-informed training of AE+NODE performs a disentanglement in the deeper layers (innermost layer) by expanding the relative volume of the high probability points, as seen in Fig. \ref{fig:PDF_LatVars}, where the ignition region is filled by more density or near uniform density to avoid the difficulty in the mode mixing. This is also manifested in Fig. \ref{fig:Latent_space} as a smooth temporal evolution of the latent variables. Thus, the stiffness reduction mechanism of the dynamics-informed training of AE+NODE becomes clearly evident from the nature of the neural network, specifically the information flow in the hidden layers and disentangling mechanism in the deeper layers.

Another point to note is that the decoder has to learn from the smooth manifold to the complex manifold. However, as discussed in the subsections \ref{Sec:IP_DPI} and \ref{Sec: Learning_Dynamics}, the learning of the decoder is from the outer layer to the inner layer. Once fully trained, the decoder maps the variable from the smooth manifold to the complex manifold. Since the decoder layers form the Markov chain, the transition from the smooth manifold to the complex manifold has to be gradual. This can be clearly seen in Fig. \ref{fig:5Layer_L5}(b) and Fig. \ref{fig:Hidden_variables}. In contrast, the encoder projects the variable from the complex manifold to the smooth manifold, which is a less complex process. Moreover, the $L_3$ loss function is mainly constraining the encoder's final layer projection to the NODE-produced trajectory as seen in Fig. \ref{fig:5Layer_L5}(a) and Fig. \ref{fig:Hidden_variables}. The methodology of constraining the projection is described in Alg. \ref{Alg2} and Fig. \ref{fig:L3_Loss}. While the present subsection describes the reason behind the mapping of complex variables to the smooth manifold for stiff dynamical systems, the following subsection will describe the effect of latent dimension on learning the stiff dynamical system in terms of a dynamics-informed learning framework.
\subsection{The effect of latent dimensions on training and prediction accuracy}\label{Sec:IP_LD}

% -------------------------------------------------------------
% Fig:: Information Plane -- 1
\begin{figure}[!t]
    \centering
    \includegraphics[width=1.0\textwidth]{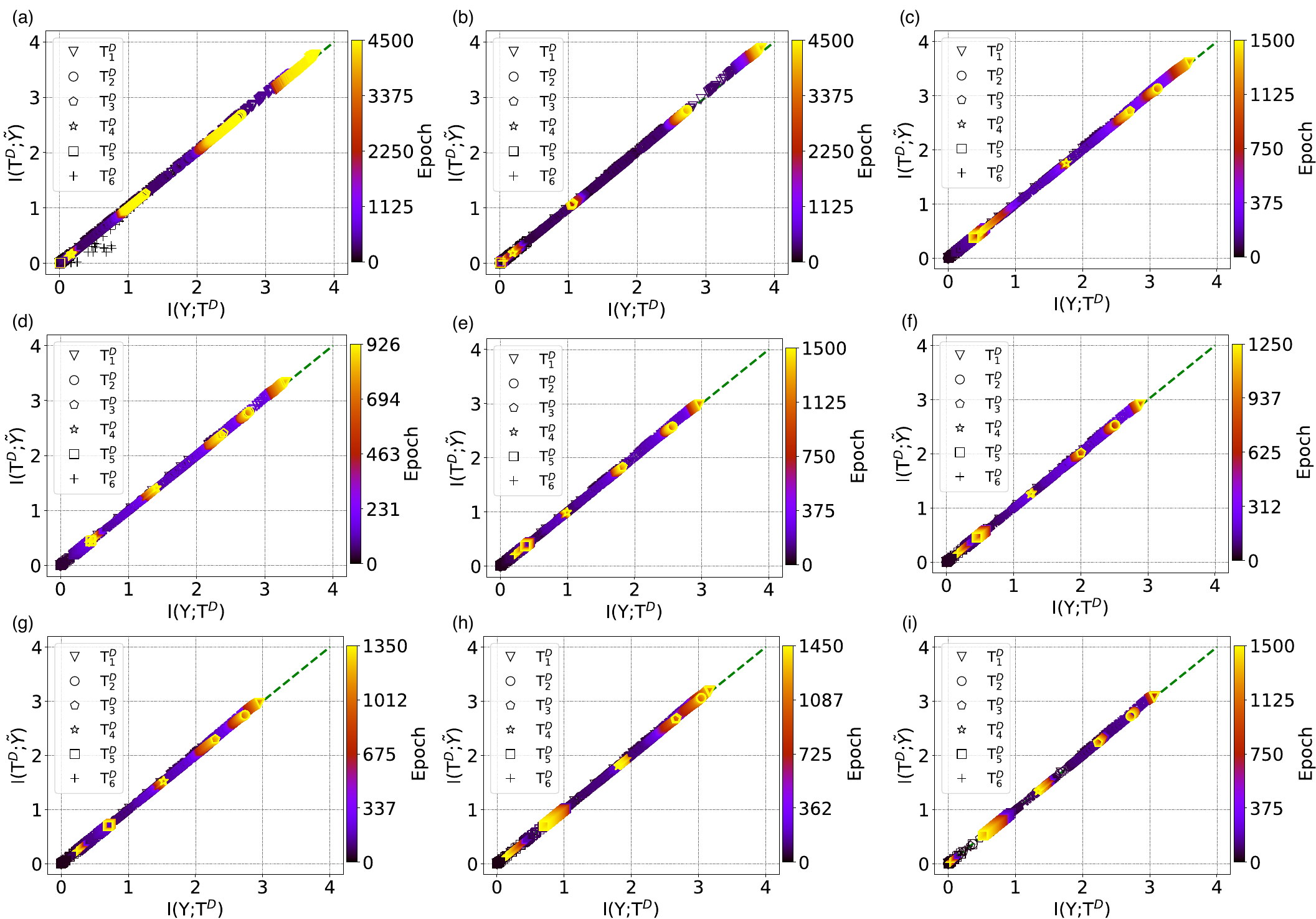}
    \caption{Information plane 1 (IP1) for the nonlinear decoder, colored with the epoch. Symbols represent the different hidden layers. Top row (a) N$_L$=1, (b) N$_L$=2, and (c) N$_L$=3; middle row (d) N$_L$=4, (e) N$_L$=6, and (f) N$_L$=7; bottom row (g) N$_L$=8, (h) N$_L$=9, and (i) N$_L$=18.}
    \label{fig:Effect_of_N_L}
\end{figure}
%Need to correct y axis label

% -------------------------------------------------------------

Lastly, it is noted that the latent space dimensions in the bottleneck play a crucial role in the ROM. While the reduced system of ODEs that evolves in latent space is easier to integrate due to the reduced number of equations,  they also play a crucial part in the accuracy of the prediction by limiting the training. Although our previous study (Ref. \cite{vijayarangan2023data}) reported the variation of accuracy of the prediction with respect to latent dimensions and hidden layers, the reason behind the variation was not fully explored. By using the information plane (IP-1), we examine the effect of the number of latent dimensions (N$_L$) on the effective AE+NODE training and accurate predictions of the chemical state space. 
 
Fig. \ref{fig:Effect_of_N_L} shows the IP-1 plane for a different number of latent dimensions (N$_L$) from 1 to 18, generated by nonlinear AE+NODE dynamic-informed training for the H$_2$-air constant pressure batch reactor. As a sanity check, it is confirmed that DPI's \ref{DPI1b} are satisfied for latent dimensions such that N$_L$ $>4$. For N$_L$ $\leq$ 3, the ($I(Y; T^D_5)$, $I(T^D_5;\tilde{Y})$) is less than ($I(Y; T^D_6)$, $I(T^D_6;\tilde{Y})$), which violates the DPI \ref{DPI1b}. In addition, the outer layer's MI ($I(Y; T^D)$, $I(T^D;\tilde{Y})$) increases in order to reduce the loss function or improve the prediction accuracy. From the error analysis values listed in Table \ref{Tab:L_H2_error}, for N$_L$ $\leq$ 3 the reconstruction error is significant. The compression phase can be observed for N$_L$ $>$ 3  in the inner layers of the decoder. Moreover, the number of latent modes in Fig. \ref{fig:Latent_space} shows the 3 distinct latent modes as observed in N$_L$ = 5. This opens up the way to find the optimum network design using  IP-1 and IP-2. Determining the optimal number of latent variables and network architecture size is an important problem that may require the study of the bifurcation phenomenon in the information plane, which is beyond the scope of the current study. It would be of interest to investigate the relation between the level of disentanglement and the underlying manifold dimensions in the context of dynamics-informed AE+NODE.

%For N$_L$ $<$ 3 (Ref. Fig. \ref{fig:Effect_of_N_L}  a, b), the state points ($I(Y; T^E)$, $I(T^E;\tilde{Y})$) in the IP-1 do not touch the limiting line (shown in dashed green line), which explains the fitting phase is not complete. However, for N$_L$ $\geq$ 3, the fitting phase is complete (Ref. Fig. \ref{fig:Effect_of_N_L} c - I). 
%From the error analysis from Table \ref{Tab:L_H2_error}, N$_L$ $<$ 3, the reconstruction error is significant. Moreover, the number of latent modes observed in the Fig. \ref{fig:Latent_space} shows the 3 distinct latent modes are observed in N$_L$ = 5 as well. 

% -------------------------------------------------------------
% Tab:: Error Latent Dimension
\begin{table}[!t]
\centering
\begin{tabular}{clllll}
\hline
\multirow{2}{*}{$N_L$} & \multicolumn{5}{c}{RRMSE [\%]} \\ \cline{2-6} 
  & T [K]     &$Y_{H_2}$     &$Y_{H_{2}O}$    &$Y_{OH}$        &$Y_{HO_2}$    \\ \hline
2       & 0.900     &6.425      &1.603      &1.783      &12.489     \\ %\hline
4       & 0.115     &0.840      &0.191      &0.300      &2.559      \\ %\hline
5       & 0.201     &1.759      &0.494      &0.420      &4.596      \\ %\hline
6       & 0.091     &0.651      &0.177      &0.254      &1.832      \\ %\hline
8       & 0.174     &1.883      &0.515      &0.433      &0.515      \\ \hline
\end{tabular}
\caption{Relative root square mean error (RRMSE) of the predicted thermochemical state from nonlinear AE+NODE (E2E) (for varying $N_L$ and fixed $H=5$) with respect to Cantera for H$_2$-air autoignition at $P=1$ atm, $\phi$=1.0, and T$_{\rm{init}}$=1000 K.}
\label{Tab:L_H2_error}
\end{table}
\section{Conclusions}\label{Sec:Conclusion}
This work demonstrates the potential of data-driven techniques to eliminate the stiffness emanating from the multitude of chemical time scales in turbulent reacting flow simulations. The effect of combining nonlinear AE with the neural ODE is studied for a constant pressure homogeneous batch reactor with H$_2$-air kinetics. The combination of nonlinear AE with NODE resulted in stiffness-removed variable evolution in the latent space. Furthermore, the predictions from dynamics-informed learning of the nonlinear AE+NODE are in good agreement with the results from direct integration with high-order implicit scheme for the ignition of H$_2$-air mixture in a constant pressure batch reactor. The rationale behind such disentangling of time scales is demonstrated using mutual information planes. 

The mutual information evolution with epochs demonstrates two distinct learning phases: i) the fitting phase and ii) the compression phase, which corroborates the IB theory. Furthermore, the nonlinear AE with $L$=5 satisfies the two DPIs, which signifies that during the compression phase, the redundant data in the dynamics are discarded, leading to an increase in information entropy. The explanation of unidirectional information flow in the autoencoder explored the rationale behind the stiffness-reduced latent manifold. The gradient propagation in the backward direction is solely responsible for the change of distribution of the latent variables concerning the physical data distribution. 

The disentangling mechanism from the better mixing hypothesis for deeper representations explained the underlying distribution of the latent space. The disentangling by deeper representations unfolds the manifold near highly probable regions and expands the relative volume of rare or low-probability events. This results in a smoother evolution of latent variables and computational gain in the integration time step. Finally, the physical significance behind the previous trial and error methodology of fixing the latent variable dimension is explained using the fitting process in the information plane. 

% -------------------------------------------------------------

% ==============================================================================
% Sec:: Acknowledgements
% ==============================================================================
\section{Acknowledgments}
This work was funded by King Abdullah University of Science and Technology (KAUST) and utilized the computational resources of the KAUST Supercomputing Laboratory (KSL).

% ==============================================================================
% Sec:: References
% ==============================================================================
%\bibliography{mybibfile}

\begin{thebibliography}{10}
\expandafter\ifx\csname url\endcsname\relax
  \def\url#1{\texttt{#1}}\fi
\expandafter\ifx\csname urlprefix\endcsname\relax\def\urlprefix{URL }\fi
\expandafter\ifx\csname href\endcsname\relax
  \def\href#1#2{#2} \def\path#1{#1}\fi

\bibitem{vijayarangan2023data}
V.~Vijayarangan, H.~A. Uranakara, S.~Barwey, R.~M. Galassi, M.~R. Malik,
  M.~Valorani, V.~Raman, H.~G. Im, A data-driven reduced-order model for stiff
  chemical kinetics using dynamics-informed training, Energy and AI (2023)
  100325.

\bibitem{steinhauser2017computational}
M.~O. Steinhauser, Computational multiscale modeling of fluids and solids,
  Springer, 2017.

\bibitem{brunton2022data}
S.~L. Brunton, J.~N. Kutz, Data-driven science and engineering: {M}achine
  learning, dynamical systems, and control, Cambridge University Press, 2022.

\bibitem{lucia2004reduced}
D.~J. Lucia, P.~S. Beran, W.~A. Silva, Reduced-order modeling: new approaches
  for computational physics, Progress in aerospace sciences 40~(1-2) (2004)
  51--117.

\bibitem{lorenz1963deterministic}
E.~N. Lorenz, Deterministic nonperiodic flow, Journal of atmospheric sciences
  20~(2) (1963) 130--141.

\bibitem{redkar2008direct}
S.~Redkar, S.~Sinha, A direct approach to order reduction of nonlinear systems
  subjected to external periodic excitations, Journal of Computational and
  Nonlinear Dynamics 3 (2008) 031011--1.

\bibitem{pavliotis2008multiscale}
G.~A. Pavliotis, A.~Stuart, Multiscale methods: averaging and homogenization,
  Vol.~53, Springer Science \& Business Media, 2008.

\bibitem{lam1989understanding}
S.-H. Lam, D.~A. Goussis, Understanding complex chemical kinetics with
  computational singular perturbation, in: Symposium (International) on
  Combustion, Vol.~22, Elsevier, 1989, pp. 931--941.

\bibitem{valorani2005chemical}
M.~Valorani, F.~Creta, D.~A. Goussis, H.~N. Najm, J.~Lee, Chemical kinetics
  mechanism simplification via {CSP}, in: 3rd MIT Conference on Computational
  Fluid and Solid Mechanics, 2005, pp. 900--904.

\bibitem{galassi2022pycsp}
R.~M. Galassi, Pycsp: A python package for the analysis and simplification of
  chemically reacting systems based on computational singular perturbation,
  Computer Physics Communications 276 (2022) 108364.

\bibitem{galassi_CSPANN}
R.~M. Galassi, P.~P. Ciottoli, M.~Valorani, H.~G. Im, An adaptive
  time-integration scheme for stiff chemistry based on computational singular
  perturbation and artificial neural networks, Journal of Computational Physics
  451 (2022) 110875.

\bibitem{bevanda2021koopman}
P.~Bevanda, S.~Sosnowski, S.~Hirche, {Koopman} operator dynamical models:
  Learning, analysis and control, Annual Reviews in Control 52 (2021) 197--212.

\bibitem{brunton2022modern}
S.~L. Brunton, M.~Budi{\v{s}}i{\'c}, E.~Kaiser, J.~N. Kutz, Modern {Koopman}
  theory for dynamical systems, SIAM Review (2022).

\bibitem{ito1998reduced}
K.~Ito, S.~S. Ravindran, A reduced-order method for simulation and control of
  fluid flows, Journal of computational physics 143~(2) (1998) 403--425.

\bibitem{ravindran2000reduced}
S.~S. Ravindran, A reduced-order approach for optimal control of fluids using
  proper orthogonal decomposition, International journal for numerical methods
  in fluids 34~(5) (2000) 425--448.

\bibitem{brunton2013reduced}
S.~L. Brunton, C.~W. Rowley, D.~R. Williams, Reduced-order unsteady aerodynamic
  models at low {Reynolds} numbers, Journal of Fluid Mechanics 724 (2013)
  203--233.

\bibitem{schmid2022dynamic}
P.~J. Schmid, Dynamic mode decomposition and its variants, Annual Review of
  Fluid Mechanics 54 (2022) 225--254.

\bibitem{malik2023dimensionality}
M.~R. Malik, R.~Khamedov, F.~E. Hern{\'a}ndez~P{\'e}rez, A.~Coussement,
  A.~Parente, H.~G. Im, Dimensionality reduction and unsupervised
  classification for high-fidelity reacting flow simulations, Proceedings of
  the Combustion Institute 39~(4) (2023) 5155--5163.

\bibitem{malik2024combined}
M.~R. Malik, R.~M. Galassi, M.~Valorani, H.~G. Im, A combined {PCA-CSP} solver
  for dimensionality and stiffness reduction in reacting flow simulations,
  Proceedings of the Combustion Institute 40~(1-4) (2024) 105532.

\bibitem{ling2016reynolds}
J.~Ling, A.~Kurzawski, J.~Templeton, {Reynolds} averaged turbulence modelling
  using deep neural networks with embedded invariance, Journal of Fluid
  Mechanics 807 (2016) 155--166.

\bibitem{zhang2020data}
J.~Zhang, W.~Ma, Data-driven discovery of governing equations for fluid
  dynamics based on molecular simulation, Journal of Fluid Mechanics 892 (2020)
  A5.

\bibitem{verma2018efficient}
S.~Verma, G.~Novati, P.~Koumoutsakos, Efficient collective swimming by
  harnessing vortices through deep reinforcement learning, Proceedings of the
  National Academy of Sciences 115~(23) (2018) 5849--5854.

\bibitem{vignon2023effective}
C.~Vignon, J.~Rabault, J.~Vasanth, F.~Alc{\'a}ntara-{\'A}vila, M.~Mortensen,
  R.~Vinuesa, Effective control of two-dimensional {Rayleigh--B{\'e}nard}
  convection: Invariant multi-agent reinforcement learning is all you need,
  Physics of Fluids 35~(6) (2023).

\bibitem{fukami2023super}
K.~Fukami, K.~Fukagata, K.~Taira, Super-resolution analysis via machine
  learning: a survey for fluid flows, Theoretical and Computational Fluid
  Dynamics 37~(4) (2023) 421--444.

\bibitem{kim2021unsupervised}
H.~Kim, J.~Kim, S.~Won, C.~Lee, Unsupervised deep learning for super-resolution
  reconstruction of turbulence, Journal of Fluid Mechanics 910 (2021) A29.

\bibitem{guemes2022super}
A.~G{\"u}emes, C.~Sanmiguel~Vila, S.~Discetti, Super-resolution generative
  adversarial networks of randomly-seeded fields, Nature Machine Intelligence
  4~(12) (2022) 1165--1173.

\bibitem{yousif2021high}
M.~Z. Yousif, L.~Yu, H.-C. Lim, High-fidelity reconstruction of turbulent flow
  from spatially limited data using enhanced super-resolution generative
  adversarial network, Physics of Fluids 33~(12) (2021).

\bibitem{vijayarangan2024reconstruction}
V.~Vijayarangan, H.~A. Uranakara, H.~G. Im, Reconstruction of high-resolution
  turbulent flow fields from sparse measurement using the diffusion normalizing
  flows, in: AIAA SCITECH 2024 Forum, 2024, p. 1362.

\bibitem{champion2019data}
K.~Champion, B.~Lusch, J.~N. Kutz, S.~L. Brunton, Data-driven discovery of
  coordinates and governing equations, Proceedings of the National Academy of
  Sciences 116~(45) (2019) 22445--22451.

\bibitem{dikemanNODE}
H.~E. Dikeman, H.~Zhang, S.~Yang, Stiffness-reduced neural {ODE} models for
  data-driven reduced-order modeling of combustion chemical kinetics, in: AIAA
  SCITECH 2022 Forum, 2022, p. 0226.

\bibitem{lee2021parameterized}
K.~Lee, E.~J. Parish, Parameterized neural ordinary differential equations:
  {A}pplications to computational physics problems, Proceedings of the Royal
  Society A 477~(2253) (2021) 20210162.

\bibitem{augustine2024survey}
M.~T. Augustine, A survey on universal approximation theorems, arXiv preprint
  arXiv:2407.12895 (2024).

\bibitem{kratsios2022universal}
A.~Kratsios, L.~Papon, Universal approximation theorems for differentiable
  geometric deep learning, Journal of Machine Learning Research 23~(196) (2022)
  1--73.

\bibitem{principe2015universal}
J.~C. Principe, B.~Chen, Universal approximation with convex optimization:
  {G}immick or reality? [discussion forum], {IEEE} Computational Intelligence
  Magazine 10~(2) (2015) 68--77.

\bibitem{liu2020selection}
H.~Liu, P.~Markowich, Selection dynamics for deep neural networks, Journal of
  Differential Equations 269~(12) (2020) 11540--11574.

\bibitem{paccolat2021geometric}
J.~Paccolat, L.~Petrini, M.~Geiger, K.~Tyloo, M.~Wyart, Geometric compression
  of invariant manifolds in neural networks, Journal of Statistical Mechanics:
  Theory and Experiment 2021~(4) (2021) 044001.

\bibitem{tishby2000information}
N.~Tishby, F.~C. Pereira, W.~Bialek, The information bottleneck method, arXiv
  preprint physics/0004057 (2000).

\bibitem{shwartz2017opening}
R.~Shwartz-Ziv, N.~Tishby, Opening the black box of deep neural networks via
  information, arXiv preprint arXiv:1703.00810 (2017).

\bibitem{saxe2019information}
A.~M. Saxe, Y.~Bansal, J.~Dapello, M.~Advani, A.~Kolchinsky, B.~D. Tracey,
  D.~D. Cox, On the information bottleneck theory of deep learning, Journal of
  Statistical Mechanics: Theory and Experiment 2019~(12) (2019) 124020.

\bibitem{noshad2019scalable}
M.~Noshad, Y.~Zeng, A.~O. Hero, Scalable mutual information estimation using
  dependence graphs, in: {ICASSP} 2019-2019 {IEEE} International Conference on
  Acoustics, Speech and Signal Processing ({ICASSP}), {IEEE}, 2019, pp.
  2962--2966.

\bibitem{geiger2021information}
B.~C. Geiger, On information plane analyses of neural network classifiers—a
  review, {IEEE} Transactions on Neural Networks and Learning Systems 33~(12)
  (2021) 7039--7051.

\bibitem{kawaguchi2023does}
K.~Kawaguchi, Z.~Deng, X.~Ji, J.~Huang, How does information bottleneck help
  deep learning?, in: International Conference on Machine Learning, PMLR, 2023,
  pp. 16049--16096.

\bibitem{alemi2016deep}
A.~A. Alemi, I.~Fischer, J.~V. Dillon, K.~Murphy, Deep variational information
  bottleneck, arXiv preprint arXiv:1612.00410 (2016).

\bibitem{voloshynovskiy2020variational}
S.~Voloshynovskiy, O.~Taran, M.~Kondah, T.~Holotyak, D.~Rezende, Variational
  information bottleneck for semi-supervised classification, Entropy 22~(9)
  (2020) 943.

\bibitem{yu2019understanding}
S.~Yu, J.~C. Principe, Understanding autoencoders with information theoretic
  concepts, Neural Networks 117 (2019) 104--123.

\bibitem{tapia2020information}
N.~I. Tapia, P.~A. Est{\'e}vez, On the information plane of autoencoders, in:
  2020 International Joint Conference on Neural Networks ({IJCNN}), {IEEE},
  2020, pp. 1--8.

\bibitem{chen2018neural}
R.~T. Chen, Y.~Rubanova, J.~Bettencourt, D.~K. Duvenaud, Neural ordinary
  differential equations, Advances in neural information processing systems 31
  (2018).

\bibitem{rubanova2019latent}
Y.~Rubanova, R.~T. Chen, D.~K. Duvenaud, Latent ordinary differential equations
  for irregularly-sampled time series, Advances in Neural Information
  Processing Systems 32 (2019).

\bibitem{paszke2019pytorch}
A.~Paszke, S.~Gross, F.~Massa, A.~Lerer, J.~Bradbury, G.~Chanan, T.~Killeen,
  Z.~Lin, N.~Gimelshein, L.~Antiga, et~al., Pytorch: An imperative style,
  high-performance deep learning library, Advances in Neural Information
  Processing Systems 32 (2019).

\bibitem{grassi2022reducing}
T.~Grassi, F.~Nauman, J.~Ramsey, S.~Bovino, G.~Picogna, B.~Ercolano, Reducing
  the complexity of chemical networks via interpretable autoencoders, Astronomy
  \& Astrophysics 668 (2022) A139.

\bibitem{kingma2014adam}
D.~P. Kingma, J.~Ba, Adam: A method for stochastic optimization, arXiv preprint
  arXiv:1412.6980 (2014).

\bibitem{zhang2020information}
Y.~Zhang, Z.~Lin, C.~K. Kwoh, Information theory-based feature selection:
  Minimum distribution similarity with removed redundancy, in: Computational
  Science--ICCS 2020: 20th International Conference, Amsterdam, The
  Netherlands, June 3--5, 2020, Proceedings, Part V 20, Springer, 2020, pp.
  3--17.

\bibitem{cover1999elements}
T.~M. Cover, Elements of information theory, John Wiley \& Sons, 1999.

\bibitem{renyi1961measures}
A.~R{\'e}nyi, On measures of entropy and information, in: Proceedings of the
  Fourth Berkeley Symposium on Mathematical Statistics and Probability, Volume
  1: Contributions to the Theory of Statistics, Vol.~4, University of
  California Press, 1961, pp. 547--562.

\bibitem{giraldo2014measures}
L.~G.~S. Giraldo, M.~Rao, J.~C. Principe, Measures of entropy from data using
  infinitely divisible kernels, {IEEE} Transactions on Information Theory
  61~(1) (2014) 535--548.

\bibitem{silverman2018density}
B.~W. Silverman, Density estimation for statistics and data analysis,
  Routledge, 2018.

\bibitem{bengio2013better}
Y.~Bengio, G.~Mesnil, Y.~Dauphin, S.~Rifai, Better mixing via deep
  representations, in: International Conference on Machine Learning, PMLR,
  2013, pp. 552--560.

\bibitem{mueller1999flow}
M.~Mueller, T.~Kim, R.~Yetter, F.~Dryer, Flow reactor studies and kinetic
  modeling of the {H}2/{O}2 reaction, International Journal of Chemical
  Kinetics 31~(2) (1999) 113--125.

\bibitem{cantera}
D.~G. Goodwin, H.~K. Moffat, I.~Schoegl, R.~L. Speth, B.~W. Weber, Cantera: An
  object-oriented software toolkit for chemical kinetics, thermodynamics, and
  transport processes, \url{https://www.cantera.org}, version 2.6.0 (2022).
\newblock \href {https://doi.org/10.5281/zenodo.6387882}
  {\path{doi:10.5281/zenodo.6387882}}.

\bibitem{tishby2015deep}
N.~Tishby, N.~Zaslavsky, Deep learning and the information bottleneck
  principle, in: 2015 {IEEE} Information Theory Workshop (itw), {IEEE}, 2015,
  pp. 1--5.

\bibitem{chang2022explaining}
S.~Chang, J.~C. Principe, Explaining deep and resnet architecture choices with
  information flow, in: 2022 International Joint Conference on Neural Networks
  (IJCNN), IEEE, 2022, pp. 1--6.

\end{thebibliography}

\end{document}